\documentclass[acmsmall,manuscript,screen]{acmart}

\AtBeginDocument{%
  \providecommand\BibTeX{{%
    \normalfont B\kern-0.5em{\scshape i\kern-0.25em b}\kern-0.8em\TeX}}}

\setcopyright{acmcopyright}
\acmJournal{TOIS}
\acmYear{2021} \acmVolume{1} \acmNumber{1} \acmArticle{1} \acmMonth{1} \acmPrice{15.00}\acmDOI{10.1145/3490519}

\acmConference[TOIS '21]{TOIS '21: ACM Transactions on Information Systems}
\acmBooktitle{TOIS '21: ACM Transactions on Information Systems}
\acmPrice{15.00}
\acmISBN{978-1-4503-XXXX-X/18/06}

\usepackage{multirow}
\newcommand{\tabincell}[2]{\begin{tabular}{@{}#1@{}}#2\end{tabular}}

\begin{document}

\title{Learning Text-Image Joint Embedding for Efficient Cross-Modal Retrieval with Deep Feature Engineering}

\author{Zhongwei Xie$^{+,*}$, Ling Liu$^{+}$, Yanzhao Wu$^{+}$}
\email{zhongweixie@gatech.edu, lingliu@cc.gatech.edu, yanzhaowu@gatech.edu}
\affiliation{%
  \institution{$^{+}$School of Computer Science, Georgia Institute of Technology}
  \city{Atlanta}
  \state{Georgia}
  \country{USA}
}

\author{Luo Zhong$^{*}$, Lin Li$^{*}$}
\email{{zhongluo, cathylilin}@whut.edu.cn}
\affiliation{%
  \institution{$^{*}$School of Computer Science and Artificial Intelligence}
  \city{Wuhan}
  \state{Hube}
  \country{China}
}
\renewcommand{\shortauthors}{Xie and Liu, et al.}

\begin{abstract}
This paper introduces a two-phase deep feature engineering framework for efficient learning of semantics enhanced joint embedding, which clearly separates the deep feature engineering in data preprocessing from training the text-image joint embedding model. We use the Recipe1M dataset for the technical description and empirical validation. In preprocessing, we perform deep feature engineering by combining deep feature engineering with semantic context features derived from raw text-image input data. We leverage LSTM to identify key terms, deep NLP models from the BERT family, TextRank, or TF-IDF to produce ranking scores for key terms before generating the vector representation for each key term by using word2vec. We leverage wideResNet50 and word2vec to extract and encode the image category semantics of food images to help semantic alignment of the learned recipe and image embeddings in the joint latent space. In joint embedding learning, we perform deep feature engineering by optimizing the batch-hard triplet loss function with soft-margin and double negative sampling, taking into account also the category-based alignment loss and discriminator-based alignment loss. Extensive experiments demonstrate that our SEJE approach with deep feature engineering significantly outperforms the state-of-the-art approaches.
\end{abstract}




\maketitle
\begin{figure}
  \centering
  \includegraphics[scale=0.22]{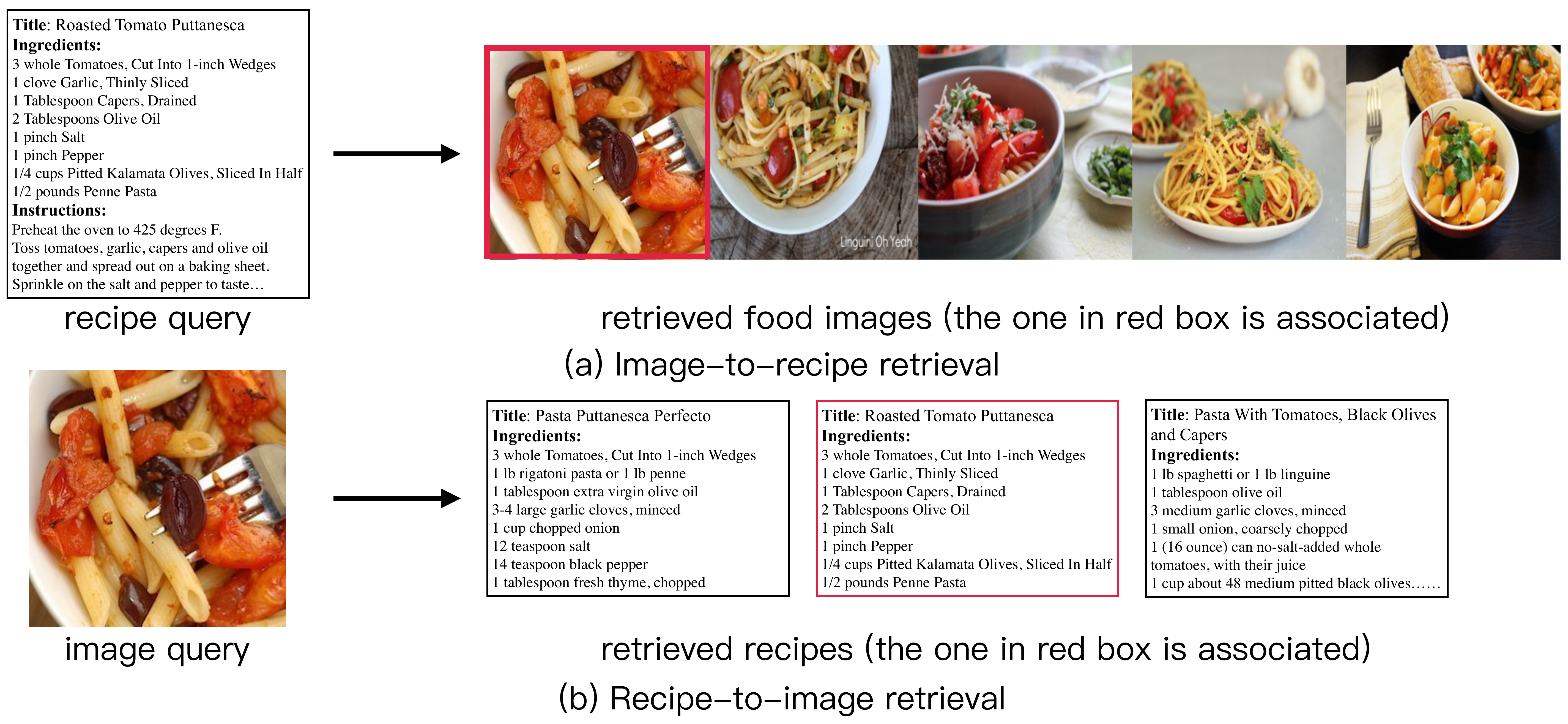}
  \caption{Examples of recipe-to-image and image-to-recipe retrieval tasks.}
  \label{retrieval-exp}
\end{figure}

\section{Introduction}
The cross-modal embedding learning problem belongs to the family of unsupervised learning~\cite{Yan+CVPR2015} and it aims to train a multi-modal joint embedding model over an unlabeled multi-modal dataset, such as Recipe1M~\cite{Salvador+CVPR2017_JESR}, by mapping features of different modalities onto the same latent space for similarity-based assessment, like cross-modal retrieval. Figure~\ref{retrieval-exp} gives examples of the recipe to image and image to recipe retrieval tasks. Most recipes provide ingredients with their quantities and cooking instructions on how ingredients are prepared and cooked (e.g., steamed or deep-fried), providing a new source of references for food intake tracking and health management. Learning cross-modal joint embeddings has been a growing area of interest for performing image to recipe and recipe to image retrieval tasks. Among the few proposals to address the cross-modal embedding problem using Recipe1M~\cite{Salvador+CVPR2017_JESR, Carvalho+SIGIR2018_AdaMine, JinJinChen+MM2018_AMSR,Hao+CVPR2019_ACME,zhu2019r2gan,lien2020recipe,fu2020mcen}, most of the recent approaches utilize the same embedding learning process to obtain the recipe and image embedding as that of~\cite{Salvador+CVPR2017_JESR}, while differ mainly in similarity metric learning to regulate the joint embeddings for the cooking recipe and food image. However, few existing approaches~\cite{lien2020recipe} have utilized the feature engineering techniques to incorporate the additional semantics in the input recipe and image data into the modality-specific embedding learning and joint embedding loss optimization.
 
In this paper, we argue that by enhancing the modality-specific embeddings by leveraging the additional semantics from the input data, we can effectively optimize the modality-specific embedding learning and regulate the cross-modal similarity loss function during iterative during the iterative joint embedding learning. For example, when learning text embedding over a dataset in the food domain, such as Recipe1M, we argue that the features from the key terms (e.g., ingredients) and the features from descriptive texts (e.g., cooking instructions) can play different roles in learning text embedding and should use different feature engineering optimizations for distinguishing its recipe from other recipes in the dataset. Similarly, the category semantics for images can be important cross-modal semantic alignment features for text and image. 

We present a novel two-phase deep feature engineering framework for efficient learning of Semantics Enhanced Joint Embedding, coined as SEJE. This paper makes three main contributions. 
{\em First}, in data preprocessing, we introduce the LSTM-based key term extraction module, term rating module and term embedding module to identify those key terms that can uniquely distinguish its recipe from other recipes and utilize the sentence embedding module to generate a vector representation for each sentence in the cooking instruction of the input recipe, prior to the LSTM-based recipe embedding learning process.   
{\em Second}, we also develop the image categorization module and category embedding module to incorporate the image category semantics into the image embedding, and iteratively align the image embedding semantically closer to the associated recipe embedding in the learned joint latent space.
{\em Third}, we propose the deep feature enhanced loss optimizations to regulate the joint embedding learning, comprising an improved batch-hard triplet loss empowered with soft-margin function and double negative sampling strategy,  category-based alignment loss and discriminator-based alignment loss, which boost the efficiency and performance of cross-modal joint embedding learning. 
Extensive experiments are conducted for cross-modal retrieval tasks on the Recipe1M benchmark dataset. The evaluation results show that empowered by the two-phase deep feature engineering techniques, our SEJE approach outperforms existing representative methods in terms of both image-to-recipe and recipe-to-image retrieval performance.

It is worth noting that although the Recipe1M dataset is used to illustrate the design of our approach and to evaluate the effectiveness of our approach compared to the state-of-the-art methods, the two-phase deep feature engineering framework is more general. For example, text feature extractions using TF-IDF, TextRank, or BERT, incorporating the image category into image embedding, and the cross-modality alignment optimizations in joint embedding are generalized methods and applicable to solving general purpose text-image cross-modal retrieval problems in other non-food domains, such as a medical procedure or health diagnosis with associate medical imaging inputs. Furthermore, Recipe1M is also a representative dataset to other domains. Most of the medical procedures or medical diagnoses tend to have a similar structured layout, such as procedure title, the list of key medical instruments and metrics used in the procedure, and the instruction for the procedure, which to a certain extent similar to recipe title, ingredients, and cooking instructions. For some social network datasets that have more missing components, such as missing the face image or missing multiple text components of a face page, the SEJE approach may need an extension to work effectively, a challenge to most of existing text-image cross-modal retrieval approaches~\cite{JinJinChen+MM2017_SAN,Salvador+CVPR2017_JESR,Carvalho+SIGIR2018_AdaMine,JinJinChen+MM2018_AMSR,Hao+CVPR2019_ACME,zhu2019r2gan,lien2020recipe,fu2020mcen}.

\section{Related Work}  
The multimedia research community is very active in investigating issues regarding food-related tasks~\cite{Joutou+ICIP2009,Kawano+2014,Yanai+2015}. Due to the development of social media, people actively post food images and recipes online, which has given rise to several important tasks based on analyzing this rich source of heterogeneous information, such as ingredient identification~\cite{Chen+2016}, recipe recommendation~\cite{Rehman+2017,Elsweiler+SIGIR-2017,Fadhil-2018} and recipe popularity prediction~\cite{Sanjo+2017}. Our focus is on another common task: food cross-modal retrieval. We aim to retrieve relevant food images or recipes in response to a recipe or food image query.

\begin{figure}[t] 
  \centering 
  \includegraphics[scale=0.24]{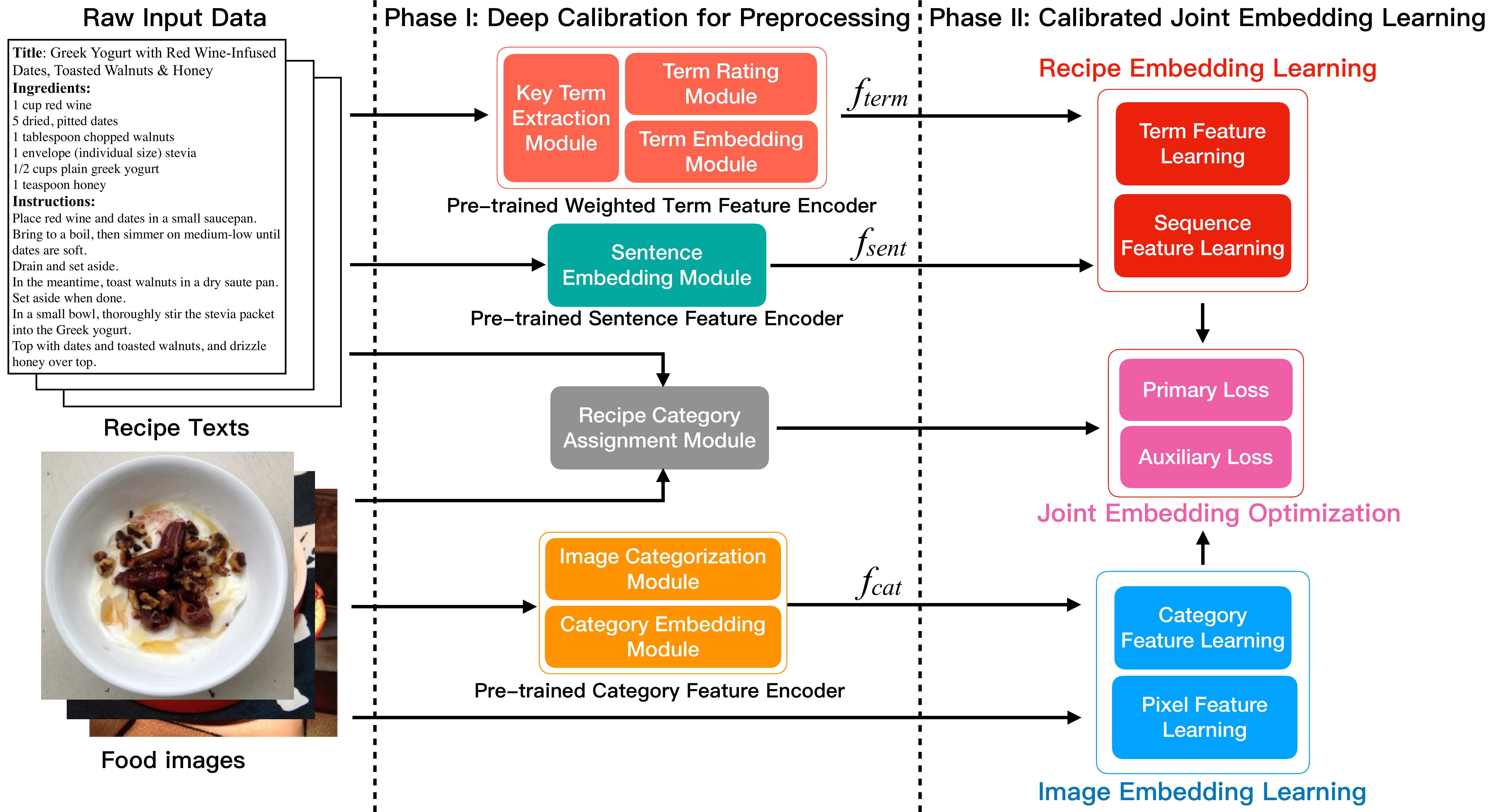}  
  \caption{The general framework of deep feature engineering for cross-modal embedding learning.}
  \label{general_framework} 
\end{figure}  

Food images are commonly diverse in terms of background content, ingredient composition, visual appearance and ambiguity. To solve this problem, early works~\cite{Jeno+SIGIR-2003,Sun-2011} have circumvented this problem by annotating images to perceive their latent semantics. However, these approaches usually require a supervision from users to annotate at least a small part of images. An unsupervised solution has emerged which consists in mapping images and texts into a shared latent space in which they can be compared. The first line of works assume a direct correspondence between visual and textual relationship and are intended to learn a similarity or distance metric from the training examples to correlate the cross-modal data. One of the classic examples is the employment of canonical correlation analysis (CCA) for semantic visual annotation~\cite{Rasiwasia+MM2010}. CCA~\cite{CCA} is one of the strongest statistic models for learning joint embeddings for different feature spaces when paired data are provided. It utilizes global alignment to allow the mapping of different modalities which are semantically similar by maximizing the correlation between associated cross-modal pairs. Most recent approaches resort to deep learning, such as deep CCA~\cite{Yan+CVPR2015}, DeViSE~\cite{Frome+2013}, correspondence auto-encoder~\cite{Feng+MM2014} and adversarial cross-modal retrieval~\cite{Wang-2017}.

Given that the problem is unsupervised learning with the raw texts of recipes, raw pixels of food images and the relation between each recipe and its associated image(s), the general trend of deep learning approaches is to learn embedding on each modality and fold each embedding into a latent space of dimension, where the embeddings of different modality can be compared directly~\cite{JinJinChen+MM2017_SAN,Salvador+CVPR2017_JESR,Carvalho+SIGIR2018_AdaMine,JinJinChen+MM2018_AMSR,Hao+CVPR2019_ACME,zhu2019r2gan,lien2020recipe,fu2020mcen}. Among the existing deep learning approaches, JESR~\cite{Salvador+CVPR2017_JESR} is a recognized pioneer from two perspectives. It introduces the first large-scale corpus of structured recipe dataset and it presents a deep neural network (DNN) based cross-modal joint embedding learning approach by leveraging the LSTM networks and CNN networks to generate the recipe embedding and image embedding respectively with pairwise cosine similarity loss and a semantic regularization term. Most of the recent methods improve JESR by upgrading the simple cosine similarity loss used for joint embedding for better learning efficiency, while keeping the same workflow for generating the recipe and image embeddings. AdaMine~\cite{Carvalho+SIGIR2018_AdaMine} extends JESR by utilizing the batch-all triplet loss with an adaptive learning scheme to eliminate zero triplets. AMSR~\cite{JinJinChen+MM2018_AMSR} and MCEN~\cite{fu2020mcen} design different attention mechanisms to map the recipe text and food image onto the joint common latent space. R$^2$GAN~\cite{zhu2019r2gan} and ACME~\cite{Hao+CVPR2019_ACME} further improve the joint embedding optimization with generative adversarial networks to learn compatible cross-modal embeddings. 

However, few of these existing methods perform additional feature engineering to extract semantic information from the raw input data and enhance the efficiency of modality-specific embedding learning. In our approach, we show that the optimization on each modality-specific embedding learning is critical for dual purposes. First, it improves the efficiency of learning modality-specific embedding, resulting in the higher quality embedding to be utilized in joint embedding learning. Second, the high-quality modality embedding can be further utilized to introduce modality alignments as additional optimizations to the joint embedding learning process, as shown in our proposed SEJE approach, a two-phase deep feature engineering framework to leveraging the additional semantics from the input text and image data, to effectively optimize both the modality-specific embedding learning and to further boost the cross-modal similarity loss function through modality alignment based regulations during the iterative joint embedding learning.

\begin{figure}[t] 
  \centering 
  \includegraphics[scale=0.19]{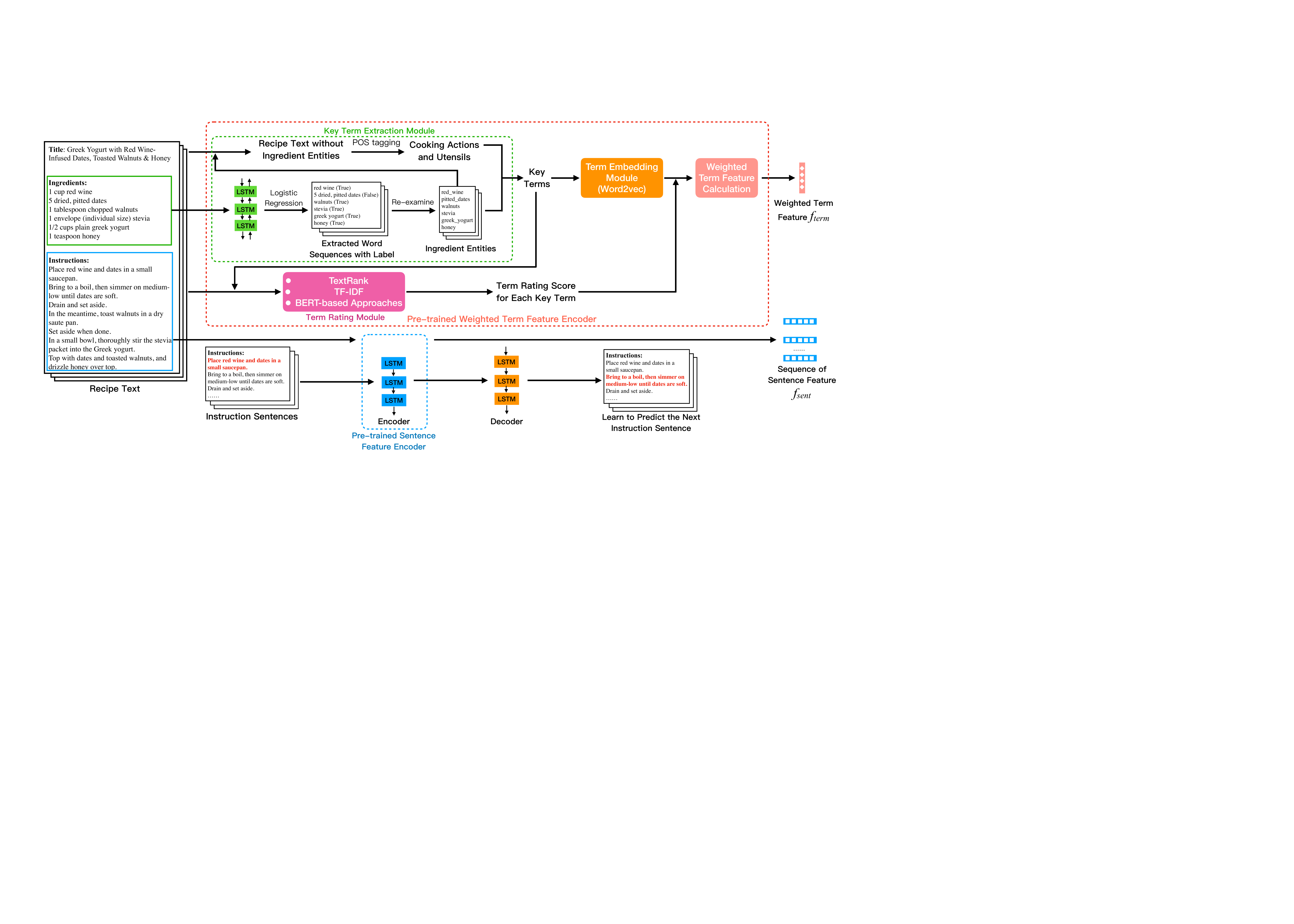}  
  \caption{The overview of the Phase I deep feature engineering for input text (recipe text) preprocessing.}
  \label{phase1-rec} 
\end{figure}

\section{Semantics Enhanced Joint Embedding}

Let $\mathbf{R}$ and $\mathbf{V}$ denote the recipe domain and image domain respectively. For a set $T$ of recipe-image pairs ($\mathbf{r}_{i}$, $\mathbf{v}_i$), where a recipe $\mathbf{r}_{i}\in \mathbf{R}$ and an image $\mathbf{v}_{i}\in \mathbf{V}$ ($1\leq i\leq T$), we want to jointly learn two embedding functions, $\mathbf{E}_{V}: \mathbf{V}\rightarrow \mathbb{R}^{d}$ and $\mathbf{E}_{R}: \mathbf{R}\rightarrow \mathbb{R}^{d}$, which encode each pair of raw recipe and food image into two $d$-dimensional vectors in the latent representation space $\mathbb{R}^{d}$.
The two embedding functions should satisfy the following condition: For $1 \leq i, j\leq T$, a recipe $\mathbf{r}_{i}$ should be closer to the matched food images $\mathbf{v}_j$ ($i=j$) than irrelevant images ($i\neq j$) in the latent $d$-dimension embedding space, and vice versa. 

We argue that in order to ensure efficient learning of the two embedding functions:  $\mathbf{E}_{R}$ and $\mathbf{E}_{V}$, we need to perform both deep feature engineering techniques in the data preprocessing phase (Phase I) and joint embedding learning phase (Phase II). Figure~\ref{general_framework} gives a sketch of our SEJE framework. In the Phase I feature engineering for preprocessing the input data, we introduce the recipe weighted term feature encoder, recipe sentence feature encoder, image category feature encoder and recipe category assignment module to preprocess the training data in three steps: (1) identifying the discriminative key terms in the recipe text, which can uniquely distinguish this recipe from other recipes in the million recipes dataset,  generating a vector representation for each key-term (such as key ingredients in a recipe); (2) generating a vector representation for each sentence in cooking instruction of the input recipe; (3) generating a vector representation of the semantic category or caption description for each image in the input data; (4) assigning each recipe-image pair with a valid category label. In the Phase II of enhanced joint embedding learning, the vectors for recipe text will be fed to the text embedding module and the vectors for the image in the same recipe will be fed together with the image pixel vector representation into the image embedding module during joint embedding training. Then deep feature enhanced loss optimization will be applied. The open-source release of SEJE is available on GitHub (https://github.com/git-disl/SEJE).

\begin{figure}[t] 
  \centering 
  \includegraphics[scale=0.23]{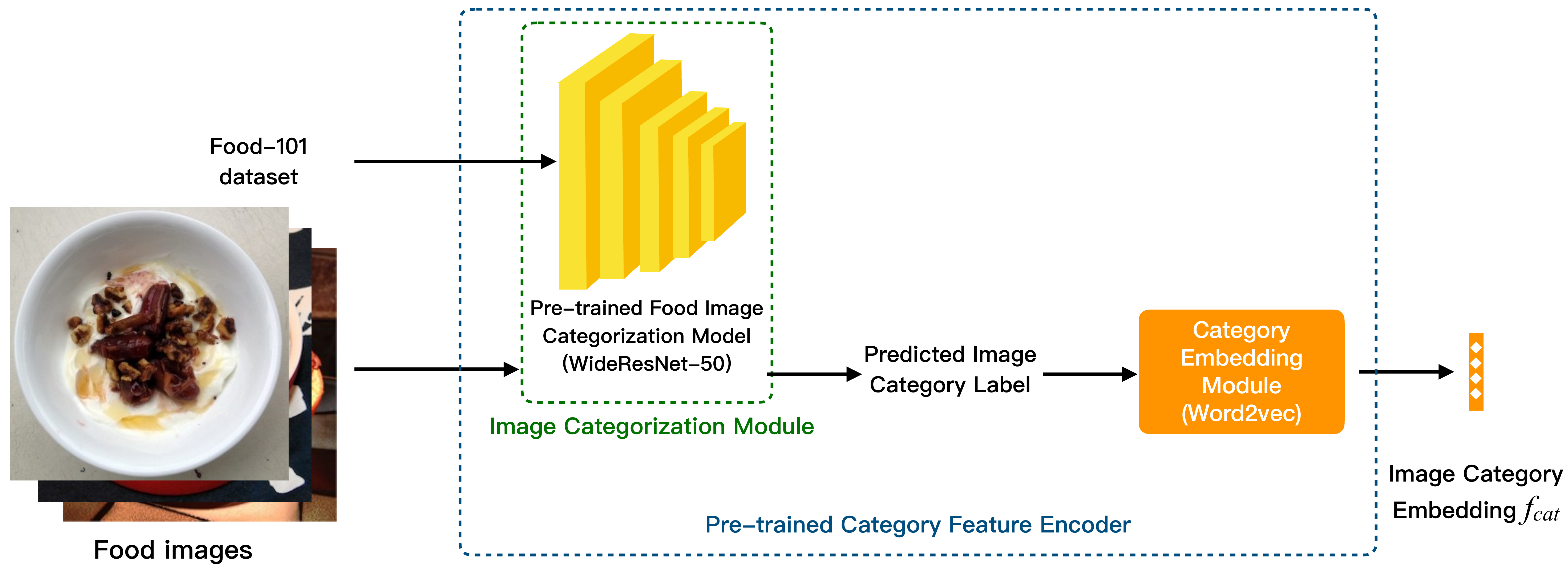}  
  \caption{The overview of the Phase I deep feature engineering for input image (food image) preprocessing.}
  \label{phase1-img} 
\end{figure}

\subsection{Deep Feature Engineering for Preprocessing}

The preprocessing phase employs a suite of deep feature engineering techniques for each modality (recipe text and food image) for dual purposes: First, it aims to enhance the quality of the enhanced text features and image features with respect to cross-modal retrieval. Second, we argue that the cross-modal joint embedding learning in Phase II can benefit significantly through semantics-enhanced deep feature engineering in Phase I. Concretely, in the preprocessing phase, we train the recipe weighted term feature encoder and the recipe sentence feature encoder to generate the weighted term feature $f_{term}$ and the sequence of instruction sentence features $f_{sent}$ from the input raw recipe text data. Figure~\ref{phase1-rec} shows the workflow for recipe text feature engineering in the input preprocessing phase. 
Similarly, we also perform deep feature engineering for food image features by training the image category feature encoder to generate the image category embedding $f_{cat}$ from the input food image. Figure~\ref{phase1-img} gives a sketch of the food image feature engineering preprocessing workflow.  Both deep enhanced text features and food image features are extracted by learning over the entire 1M recipes dataset and will be utilized for joint embedding learning through deep feature engineering of cross-modal alignments. In addition, we design a recipe category assignment module to generate the category label for each input recipe-image pair in the dataset, which will also be an input feature for joint embedding learning.

\subsubsection{Recipe Sentence Feature Encoder} 
Since the recipe instructions comprise multiple sentences, simply using a single LSTM model to learn the overall embedding from the lengthy instructions is not well suited. It is because the gradients could be diminished over so many time steps during the LSTM network training. Hence, we pre-train a sentence feature encoder to generate the representation for each instruction sentence in the data preprocessing phase, and then an LSTM model is trained over the generated sequence of instruction sentence features to learn the overall recipe sequence feature as a part of the recipe embedding learning during the joint embedding learning phase. To get the sentence feature encoder, we need to train a sequence-to-sequence model over the recipe instructions in the dataset based on the LSTM networks and skip-thoughts~\cite{Kiro-2015}, which encodes an instruction sentence and predicts the previous and next sentences according to that encoding as context information. The encoder part in the trained sequence-to-sequence model is utilized as the recipe instruction sentence encoder to generate the representation for each recipe instruction sentence in our data preprocessing phase.

\subsubsection{Recipe Weighted Term Feature Encoder} 
The common problem in the LSTM-based existing works~\cite{Carvalho+SIGIR2018_AdaMine,JinJinChen+MM2017_SAN,Salvador+CVPR2017_JESR,Hao+CVPR2019_ACME,zhu2019r2gan,lien2020recipe} is that each key term (ingredients) are treated equally, without noticing the fact that the discrimination significance of each key term for its recipe is different (e.g. the frequently appeared ingredients like water and oil have less discriminative ability to distinguish its recipe from other recipes). The recipe weighted term feature encoder is utilized to capture the discriminative semantics of each key term in the data preprocessing phase and help improve the quality of learned recipe embedding in the joint embedding phase by correcting the potential errors in the sequence feature engineering, which consists of three core components: key term extraction module, term rating module and term embedding module.

 \begin{figure*}[tbp]
  \centering
  \includegraphics[scale=0.4]{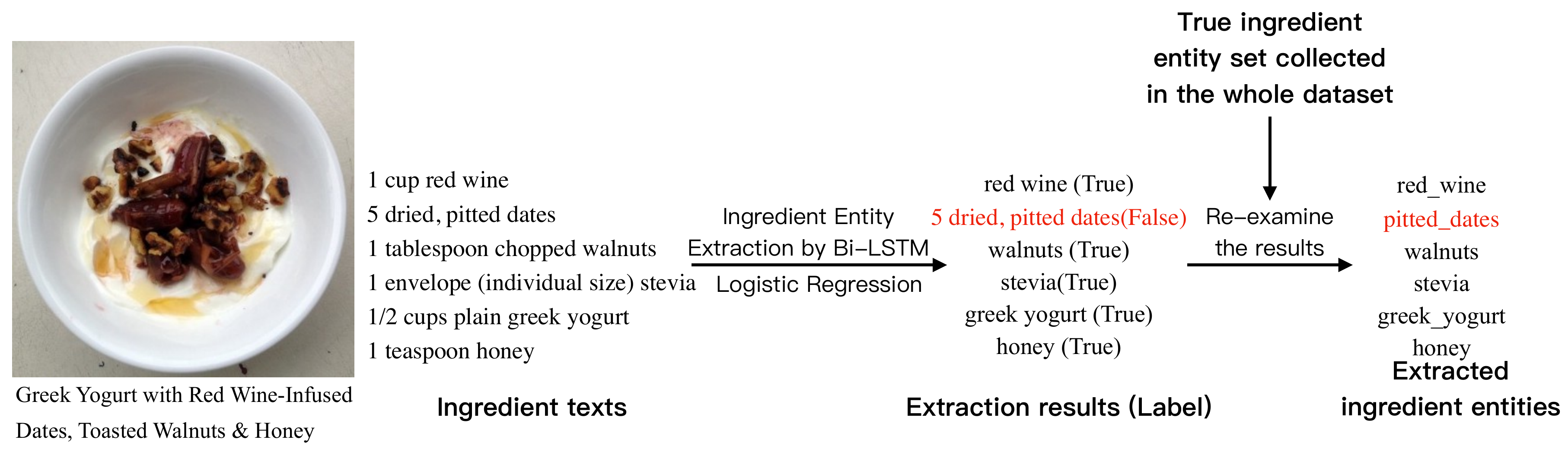}
  \caption{ The workflow of extracting the ingredient entities is depicted in the red dashed box using an example recipe and the matched food image. SEJE utilizes deep feature engineering in preprocessing phase to refine the LSTM extracted key ingredient results. In this example, LSTM classifies the term ``pitted\_dates'' as the false result in terms of a key ingredient. In SEJE, we first perform LSTM based term extraction and NLP based term extraction and ranking independently using BERT, TFIDF, or TextRank, and then combine the two independent and yet complementary term extraction and ranking methods to produce the final set of key ingredients, effectively correcting the errors from the LSTM key ingredient extraction.}
\label{miss_ingr} 
\end{figure*}

\paragraph{\bf Key term extraction module} Ingredient entities (e.g., potato), cooking utensils (e.g., oven) and actions (e.g., blend) are the three main categories of key terms extracted in our data preprocessing phase, as illustrated in the top left part in the Figure~\ref{phase1-rec}. Concerning the ingredient entity extraction, a bi-directional LSTM network is trained over the ingredient texts in the dataset to analyze the word sequence of each ingredient text and learn to identify those word sequences with high probability as ingredient entity, like the word sequence ``red wine" from the ingredient text ``1 cup red wine". Due to the accuracy limitation of the LSTM-based ingredient entity extractor, the binary logistic regression is performed on each extracted word sequence to verify it is an ingredient entity or not. Those word sequences with true labels are regarded as the ingredient entities in SEJE. But those with false labels still might contain the ingredient entities.
Rather than bluntly discarding those false word sequences in the existing works, we re-examine each false word sequence to see if there exists any true ingredient entity that appears in other recipe ingredient texts in the whole dataset. If yes, we will also regard this ingredient entity from the false word sequence to make sure that we do not miss any true ingredient entity. Figure~\ref{miss_ingr} shows an illustration of using the true ingredient entity set collected in the whole dataset to re-examine those word sequences marked with false labels and effectively recover the important ingredient entity ``pitted\_dates" that would be discarded in the existing works. As for the extraction of cooking utensils and actions, we first process the recipe title and instructions by removing the ingredient entities identified in its ingredient text and then leverage the part-of-speech tagging techniques~\cite{kumawat2015pos} to extract the nouns and verbs in the remaining text as the cooking utensils and actions.

\paragraph{\bf Term rating module} Another advantage of introducing NLP-based term extraction using BERT, TF-IDF, or TextRank is to utilize the term ranking results produced during their term extraction process. For example, each key term may have different discrimination significance with respect to distinguishing its recipe from other recipes in the entire Recipe1M dataset (e.g., the ingredient ``artichoke'' is more unique for identifying its recipe, compared to some other ingredients, such as ``salt''). Hence, in SEJE, each key term is assigned with a significance weight using the term ranking score provided by the respective term rating algorithm, such as TF-IDF~\cite{Salton-1988}, TextRank~\cite{mihalcea2004_textrank} or BERT~\cite{devlin2018_bert} based approach. Prior to performing the key term extraction and computing their ranking scores using one of the NLP-based algorithms, we first using bi-directional LSTM to learn those key ingredient entities in the recipe text that are word sequences instead of single words. This step enables important word sequences to be extracted as ingredient entities by treating each word sequence as a single key term with words connected using underline, like ``pitted\_dates". For the term extraction and ranking algorithms focusing on the occurrence frequency of each term, like TF-IDF and TextRank, the ranking score for each key term can be obtained by performing the algorithm on the entire set of recipe texts. For the other type of term extraction and ranking algorithms that utilize the semantic correlation between each term and its recipe text, like the BERT-based approach, the representations of each key term and its corresponding recipe text are obtained first using the specific pre-trained BERT model, and then the term ranking score is computed based on the cosine similarity score between the representations of each key term and its corresponding recipe text. In SEJE Phase I feature engineering, the average time for performing the term extraction and ranking in one recipe by using TF-IDF, TextRank, BERT, DistilBERT and RoBERTa are 33.3ms, 1.1ms, 62.8ms, 34.5ms and 60.4ms respectively.

\paragraph{\bf Term embedding module} Word2vec model~\cite{Mikolov-2013} with the Continuous Bags of Words (CBOW) model framework is utilized in SEJE to generate the embedding for each key term. Similar to the pipeline in the term rating module, the recipe texts need to be processed first to connect the word sequences of the ingredient entities with underline before start training the word2vec model over the recipe texts in the dataset. Once the training process of the word2vec model is complete, every key term in the recipe corpus can get its word2vec representation through the learned word embedding matrix, which captures the word sequence patterns in the recipe texts. Finally, the recipe weighted term feature can be generated by combining the representations of all the key terms from the term embedding module weighted by their significance weight values from the term rating module. 

 \begin{figure}[t] 
  \centering 
  \includegraphics[scale=0.2]{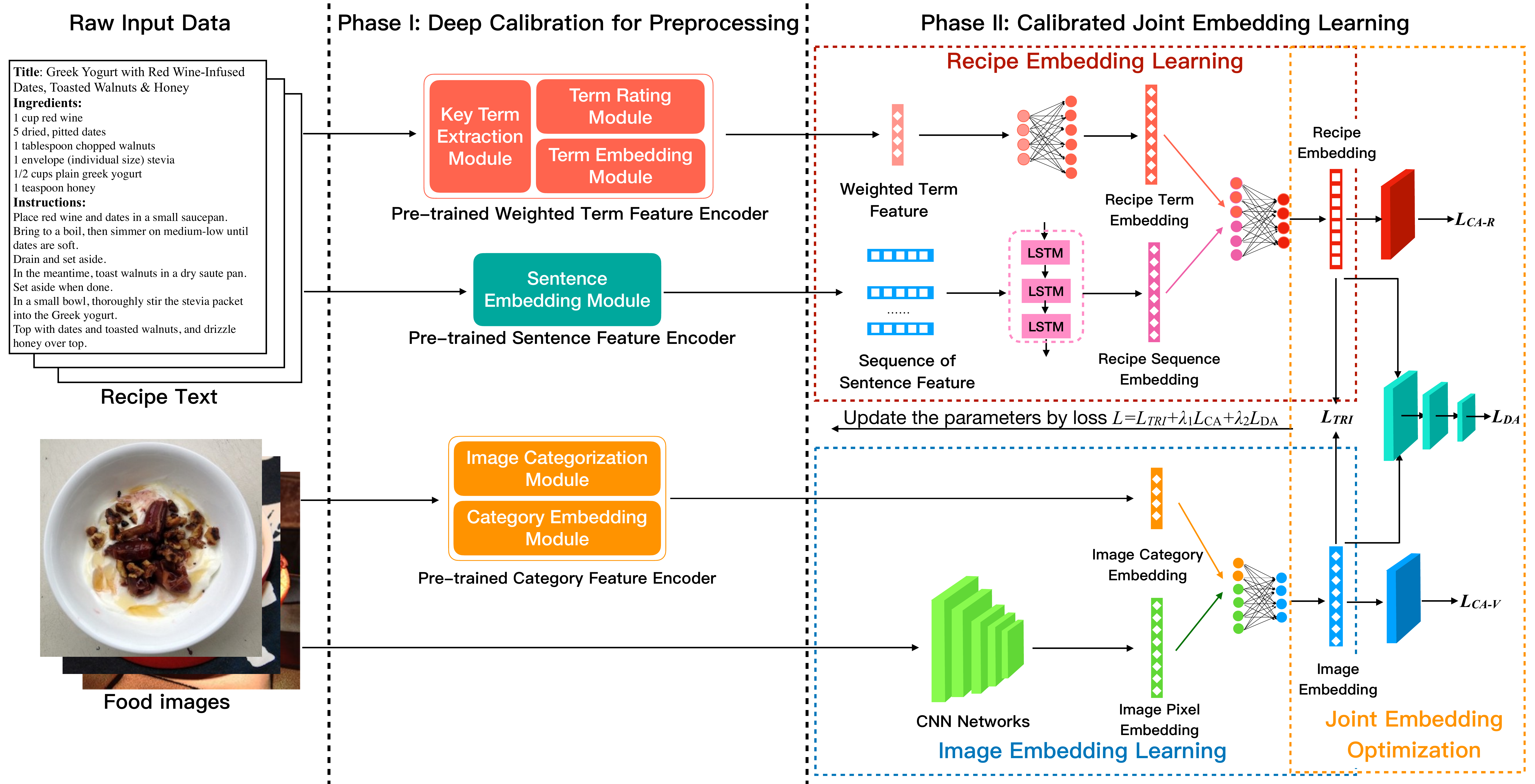}  
  \caption{The overview of the phase II enhanced joint embedding learning.}
  \label{phase2} 
\end{figure}  

\subsubsection{Image Category Feature Encoder} 
Almost all the existing works learn the food image embedding by simply leveraging the pre-trained CNN networks as image encoder, e.g., VGG-16~\cite{Simonyan-2014} and ResNet-50~\cite{He+CVPR2016}, overlooking the fact that the food images corresponding to a cooking recipe may have different visual appearances due to different ways of using ingredients, different decorations and background. To deal with this issue, we build an image category feature encoder to leverage the category semantics feature from the input food image to help iteratively align the learned image embedding closer to the associated recipe text embedding in the common latent space during the joint embedding learning phase. The image category feature encoder comprises two components: the image categorization module and category embedding module. For the image categorization module, we resort to the Food-101 dataset~\cite{Food101} and leverage its image category knowledge information by training an image category classifier over the food images in the Food-101 dataset, based on the WideResNet-50 model~\cite{wideResnet}, which can predict one of the Food-101 categories given a food image. And for the category embedding module, we also utilize the word2vec model which is trained in the recipe weighted term feature encoder. Given a food image, a food category can be obtained by the image categorization module and then the category embedding module can generate the image category embedding according to the predicted food category label, as illustrated in Figure~\ref{phase1-img}.

\subsubsection{Recipe Category Assignment Module} 
The category label for the recipe-image pairs in the Recipe1M dataset is first generated by the work of JESR~\cite{Salvador+CVPR2017_JESR}, which are directly utilized by the following works~\cite{Carvalho+SIGIR2018_AdaMine,JinJinChen+MM2018_AMSR,Hao+CVPR2019_ACME,zhu2019r2gan,lien2020recipe}. However, this category assignment algorithm is only obtained from the frequent bigrams of the recipe titles in the Recipe1M dataset and category labels of Food-101 dataset, resulting in nearly half of recipe-image pairs in the Recipe1M dataset without a valid category label. Hence, the category assignment algorithm has to be optimized to improve its effectiveness. In SEJE, the category labels in the Food-101 dataset are regarded as more reliable than the frequent bigram results from the recipe titles, since the labels in the Food-101 dataset are manually designed, while JESR values frequent bigram results more. First, for a recipe whose title contains one of the Food-101 labels, we will assign this category label to this recipe-image pair. Second, we will rank the bigrams obtained from the recipe titles by the occurrence frequency and the bigrams whose frequency is lower than a threshold will be eliminated (The threshold is set as 25 in SEJE). For a bigram from the highest frequency to the lowest, if a recipe in the remaining recipes without labels after the first step contains this bigram, its recipe-image pair will be assigned with this bigram as the category label. Third, if a recipe without label whose ingredient or instruction texts contain one of the Food-101 labels and obtained bigrams, its recipe-image pair will be labeled with this category. Finally, for the remaining recipe-image pairs still without labels, the top-1 category label results of the food images predicted by using the aforementioned image categorization module will be assigned to their recipe-image pairs as category labels. Therefore, every recipe-image pair in SEJE can be assigned a valid category label from the recipe category assignment module.

\subsection{Enhanced Joint Embedding Learning} 
After finishing the Phase I feature engineering to preprocess the raw training data, the output will be fed into the Phase II deep feature engineering for joint embedding learning to train a joint embedding model over the whole recipe dataset, as illustrated in Figure~\ref{phase2}. The joint embedding model learns the recipe text embedding, the image embedding, and the joint embedding loss function concurrently by taking one recipe training data at a time. The sequence of sentence embeddings for recipe instructions and the recipe weighted term feature from the Phase I feature engineering in the data preprocessing are the input features for the recipe embedding learning. An LSTM model is trained over the sequence of instruction sentence features to generate the overall instruction embedding. And the recipe weighted term feature is passed through a fully-connected layer to obtain the recipe term embedding. By performing the concatenation operation on the recipe term embedding and recipe sequence embedding, the recipe embedding can be produced followed by a fully-connected layer. As for the image embedding learning, ResNeXt-101 model~\cite{xie2017aggregated} is utilized to learn the image pixel embedding, which can generate the final image embedding together with the image category feature from the Phase I feature engineering by feeding their concatenation result into a fully-connected layer. And the learned recipe and image embedding will be optimized by the joint embedding loss function in the Phase II feature engineering.

\subsubsection{Joint Embedding Loss Optimization} 

To efficiently optimize the joint embedding learning, we adopt the soft-margin based batch-hard triplet loss empowered with a novel double-negative sampling strategy as the primary loss, denoted as $L_{TRI}$, together with two auxiliary loss regularizations to facilitate the cross-modal joint embedding learning, which are category-based alignment loss on both cooking recipe and food image $L_{CA}$ and discriminator based alignment loss $L_{DA}$. The overall objective function is defined as follows: 
\begin{equation} 
L = L_{TRI} + \lambda_1 L_{CA} + \lambda_2  L_{DA} 
\end{equation} 
\noindent  
where $\lambda_1$ and $\lambda_2$ both are trade-off hyper-parameters. We leverage cross-validation to configure these parameters empirically. By experimentally evaluating their impacts through cross-validation by varying the $\lambda_1$ and $\lambda_2$ from 0.001 to 0.02, it is observed that the best results can be obtained when $\lambda_1$ and $\lambda_2$ both are set as 0.005.

 \begin{figure}
  \centering   
  \includegraphics[scale=0.3]{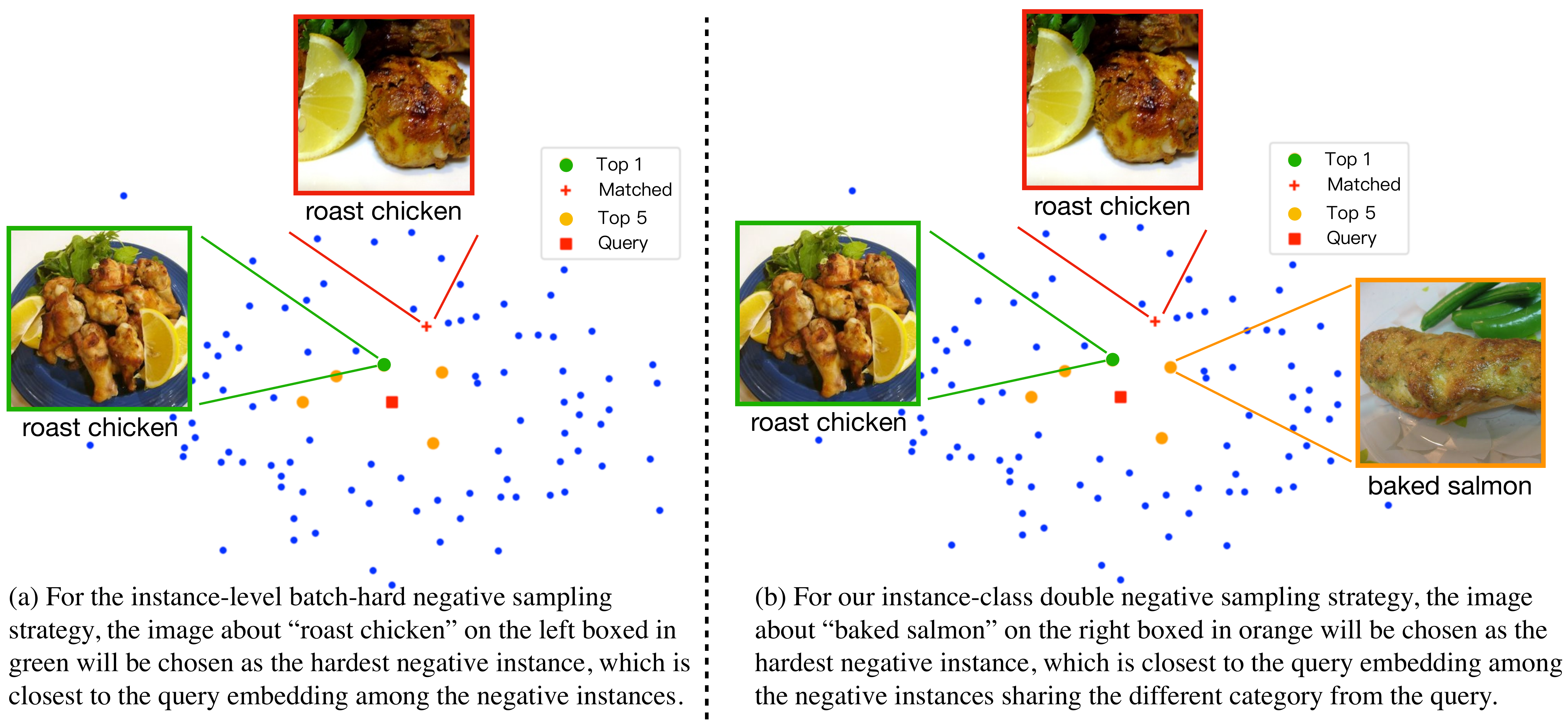}   
  \caption{ Illustration of our instance-class double hard sampling strategy.}
  \label{negative_sample}   
\end{figure}

\paragraph{\bf Double negative sampling and soft-margin optimized batch hard triplet loss}
During the training stage, triplet loss can be computed given a triplet $(x_a, x_p, x_n)$ among the training samples, where $x_a$ denotes the embedding for the anchor point in one modality,  $x_p$ and $x_n$ represent the positive and negative embeddings from the other modality respectively. The triplet loss aims to minimize the distance between the anchor point and positive instance, and maximize the distance between the anchor point and negative instance. By selecting the hardest positive and negative samples for each anchor point within every batch when calculating the triplet loss, Hermans~\cite{Hermans+2017} demonstrates that it often outperforms the {\em batch-all triplet loss}, which is calculated on all possible triplets in the training set. In SEJE, the batch-hard triplet loss is optimized by introducing a double hard negative sampling strategy. We define two types of batch hard positive or negative samples, i.e., instance-level and class-level batch hard samples. For instance-level batch hard sample, given an anchor image (or recipe) embedding (i.e., $\in \mathbb{R}^d$) in each batch, there is only one positive recipe (or image) instance in the batch, which corresponds to this anchor image (or recipe). So it will be the instance-level batch hardest positive sample and the rest of the recipe (or image) instances in the batch will be considered as negative examples of the anchor image (or recipe). The instance-level batch hardest sample is defined as the one whose vector distance to the anchor point embedding is the smallest among the negative samples in the batch. As for the class-level batch hard sample, those recipe (or image) instances that have the same category as the anchor image (or recipe) are considered as positive samples, while the rest with different categories are regards as negative samples for this anchor. We define the class-level batch hardest positive or negative recipe (or image) sample as the one whose vector distance to the embedding of anchor image (or recipe) is the largest or smallest among all the class-level positive or negative samples in the batch respectively. \noindent Figure~\ref{negative_sample} illustrates the effectiveness of our instance-class double sampling strategy used in our $L_{TRI}$ alignment loss optimizer. 
When only using the instance-level batch hard strategy, given a recipe query (denoted as the red square), the images in the red and green boxes will be the hardest positive and negative instances respectively, since the image boxed in red is the closest among the negative image instances. However, this instance-level batch hardest negative instance shares the same category ``roast chicken'' as the anchor recipe query. In comparison, the instance-class level batch hardest negative sample will be the ``baked salmon'' image in the orange box, which is the next closest negative instance to the anchor recipe. Therefore, given a recipe query about ``roast chicken'', SEJE is more likely to retrieve the images about ``roast chicken", rather than those about ``baked salmon''.

Besides, the softplus function $ln(1+\exp(\gamma(\cdot+m)))$ is used as a smooth approximation in SEJE to replace the hinge function $[m+\cdot]_+$ used in existing works~\cite{Carvalho+SIGIR2018_AdaMine,JinJinChen+MM2018_AMSR,Hao+CVPR2019_ACME}, which assumes that the distance between the anchor point and negative instance is always larger than the distance between the anchor and positive instance by a fixed margin $m$. Our soft-margin based approach improves the hinge with an exponential decay instead of a threshold-based hard cut-off. 
The soft-margin based batch-hard triplet loss with a double sampling strategy, denoted by $L_{TRI}$, is given as:

\begin{equation}  
\label{DHTL}  
\begin{aligned}  
L_{TRI}  = \sum^{N}_{i=1}ln(1+e^{\gamma({d(E_{r_i}^a, E_{v_i}^p)}-\min{d(E_{r_i}^a, E_{v_i}^n)} + m)})  + \sum^{N}_{i=1}ln(1+e^{\gamma( d(E_{v_i}^a, E_{r_i}^p)-\min d(E_{v_i}^a, E_{r_i}^n) + m)} )  
 \end{aligned}  
 \end{equation} 
 
\noindent  
where $d(\cdot)$ measures the Euclidean distance between two input vectors, $N$ is the number of the different recipe-image pairs in a batch, subscripts $a$, $p$ and $n$ refer to anchor, positive and negative instances respectively, $E_{r_i}, E_{v_i}$ refer to the embeddings of the recipe and image in the $i$-th recipe-image pair respectively, $\gamma$ is the scaling factor and $m$ denotes the margin of error in the triplet loss.

\paragraph{\bf Category-based alignment loss}

The category-based loss regularizations on both learned recipe and image embeddings are utilized in SEJE to further optimize the joint embedding learning, which require the learned matched recipe and image embedding should contain the same category semantics information by reducing the cross-entropy loss between the modality-specific embedding and the corresponding category (recall $L_{CA-R}$ and $L_{CA-V}$ in Figure~\ref{general_framework}).  
SEJE assigns every recipe-image pair to one of the 1005 category labels learned from text mining analysis on the whole recipe text and the image categorization module in the Phase I deep feature engineering, avoiding assigning ``background'' labels to a large percentage of recipe-image pairs as done in existing approaches~\cite{Salvador+CVPR2017_JESR,Carvalho+SIGIR2018_AdaMine,JinJinChen+MM2018_AMSR,Hao+CVPR2019_ACME}. 
We utilize the cross-modal category distribution alignment between the textual recipe and visual image as a regularization to our primary joint embedding loss $L_{TRI}$. The category-based loss regularization is applied to both image and recipe embeddings as follows:

\begin{equation}  
L_{CA-R} = -\sum^{N}_{i=1}\sum^{N_c}_{t=1}{y^{i,t}_R \log(\hat{y}^{i,t}_R)}  
 \end{equation} 
 
 \begin{equation}  
L_{CA-V} = -\sum^{N}_{i=1}\sum^{N_c}_{t=1}{y^{i,t}_V \log(\hat{y}^{i,t}_V)}  
 \end{equation} 

\noindent   
where $L_{CA-R}$ is the loss of regularization on the recipe embedding, while $L_{CA-V}$ is on the image embedding. $N_c$ is the number of category labels, $N$ is the number of the different recipe-image pairs in a batch, $y^{i,t}_R$ and $\hat{y}^{i,t}_R$ are the true and estimated possibilities that the $i^{th}$ recipe embedding belongs to the $t^{th}$ category label, and similarly, $y^{i,t}_{V}$ and $\hat{y}^{i,t}_{V}$ are defined for image embedding.

\begin{figure}
  \centering   
  \includegraphics[scale=0.21]{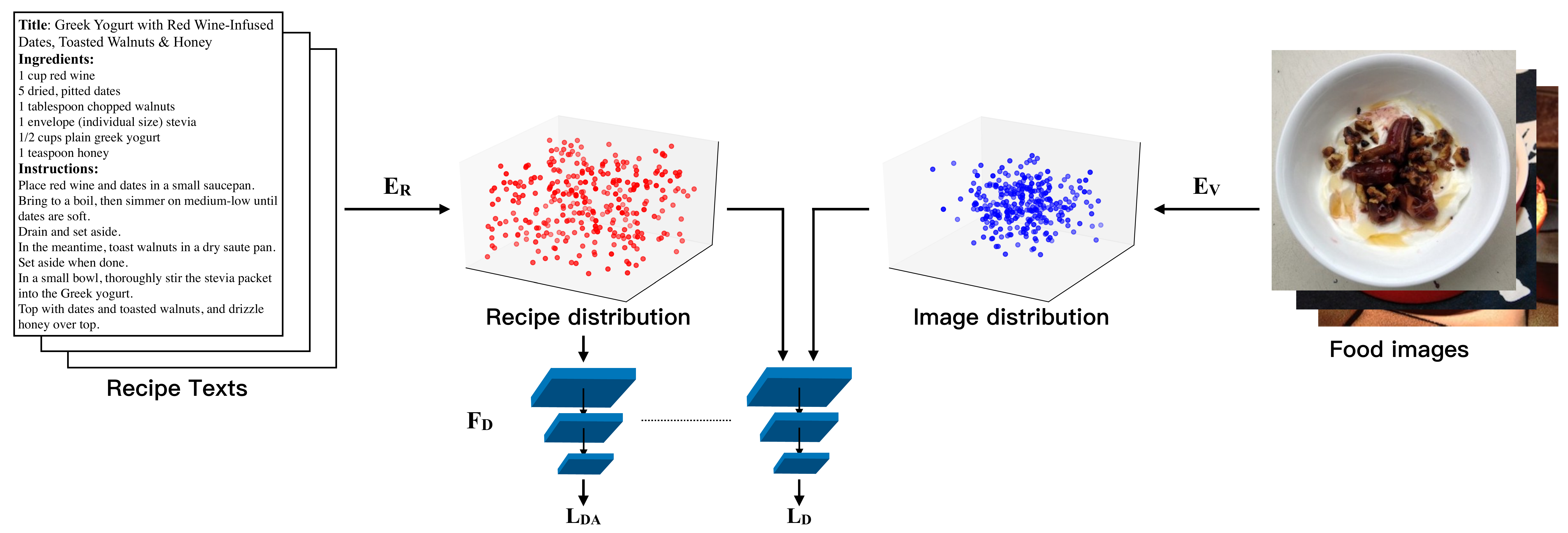}   
  \caption{ Illustration of discriminator based alignment loss regularization.} 
  \label{discriminator_nets}   
\end{figure}

\paragraph{\bf Discriminator based alignment loss}
In order to yield good performance of joint embedding learning, aligning the distributions of the two heterogeneous modalities (i.e., recipe text and food image) should be taken into consideration. We achieve this by utilizing the competing strategy in GAN~\cite{goodfellow+2014} with the gradient penalty~\cite{gulrajani+2017} to further regularize the joint embedding loss optimization. During the learning of joint embeddings, we train a discriminator model in the meantime such that for each pair of matched recipe and image, given an embedding from one modality, this discriminator can tell the modality source of the embedding, which is against our goal of joint embedding learning. When the trained discriminator cannot easily discriminate the embedding of one modality from the other, it can be considered that the learned distributions of cross-modal embeddings are too similar in the common latent space $\mathbb{R}^d$ for the discriminator to tell them apart. Since the recipe and image embeddings play the identical role during the discriminator learning, we set the learned discriminator as the one predicting the confidence value of the input embedding from the image modality (i.e., when receiving the image embedding, this discriminator will give a high confidence value while recipe embedding will result in low confidence value). Therefore, the higher confidence values our discriminator gives when receiving the recipe embeddings, it indicates that the more aligned and similar the distributions of learned recipe embeddings and image embeddings are in the latent common space $\mathbb{R}^d$. The illustration of how the discriminator works is showed in Figure~\ref{discriminator_nets}. We below define the loss $L_D$ for the training of our discriminator model, which is made up of three fully-connected layers and the cross-modal discriminator based alignment loss regularization $L_{DA}$ respectively:

\begin{equation} 
\begin{aligned} 
L_D&=\sum^{B}_{i=1}[\log(F_D(E_R(r_i)))+\log(1-F_D(E_V(v_i)))+\lambda_D(\Vert \nabla_{x_i}\log(F_D(x_i))\Vert_2-1)^2] 
 \end{aligned} 
 \end{equation} 
\begin{equation} 
L_{DA} = \sum^{B}_{i=1}{\log(1-F_D(E_R(r_i)))}  
\end{equation} 

\noindent 
where $F_D(\cdot)$ is the function of our trained discriminator, which outputs the confidence value of how confident it is to predict the modality source of the input embedding is image, $B$ is the number of the different recipe-image pairs in a batch, $\lambda_D$ is the trade-off parameter and set to 10 as suggested in~\cite{gulrajani+2017}, $x_i$ is a random interpolation between the $i^{th}$ recipe embedding $E_R(r_i)$ and image embedding $E_V(v_i)$.

\section{Experiments}
\label{experiments}

\subsection{Dataset and Evaluation Metrics.}
We evaluate the effectiveness of different approaches on the Recipe1M dataset, consisting of over 800K recipes (title, list of ingredients and cooking instructions) and 1 million associated food images. Following the preprocessing in~\cite{Salvador+CVPR2017_JESR}, duplicate recipes, recipes without images, the unreadable recipes without any nouns or verbs are filtered out, resulting in 238,399 matching pairs of images and recipes for the training set, and 51,116 and 51,303 matching pairs for validation and test respectively.
Experiment setup follows the existing approaches: (i) Randomly sample 10 unique subsets of 1k or 10k matching recipe-image pairs from the test set. (ii) Use each item in one modality as a query (e.g., an image), and rank instances in the other modality (e.g., recipes) by the Euclidean distance between the query embedding and each candidate embedding from the other modality in the test set. (iii) Calculate the median retrieval rank (MedR) and the recall percentage at top K (R@K), i.e., the percentage of queries for which the matching answer is included in the top $K$ results ($K$=1,5,10). We report the average metrics on the 10 groups of randomly chosen samples. Note that even though many loss regularizations are added in our framework, these optimizations are only performed at the training stage but not in the testing stage. 

\paragraph{\bf Implementation Details.} The three types of BERT models used in our term rating module are the publicly released pre-trained models\footnote{https://www.sbert.net/docs/pretrained\_models.html}. The POS tagging used in this work is implemented by using the public Python library TextBlob~\cite{loria2018textblob}. The word2vec model training is implemented by using the code provided on the Google Code Project Hosting Service\footnote{https://code.google.com/p/word2vec/}. The dimensions of joint embedding and word2vec embedding are set as 1024 and 300 respectively. Adam optimizer~\cite{Adam-2014} is employed for model training with the initial learning rate set as $10^{-4}$ in all experiments, with the mini bath size of 100. All deep neural networks are implemented on the Pytorch platform and trained on a single Nvidia Titan X Pascal server with 12GB of memory.

 \begin{table} 
		\center 
		\caption{{Performance comparison of our SEJE with existing representative methods on the 1k and 10k test set. The symbol ``-'' indicates that the results are not available from the corresponding works. The symbol ``*'' in the AdaMine$^*$ and ACME$^*$ indicate that their results are evaluated by us based on their provided pre-trained models.}} 
		\label{main_results} 
		\begin{tabular}{cc|cccc|cccc} 
		\hline
		\multirow{2} * {\tabincell{c}{Size of \\test-set}} & \multirow{2} * {Approaches}  &\multicolumn{4}{c}{Image to recipe retrieval} & \multicolumn{4}{c}{Recipe to image retrieval } \\ 
		 \cline{3-10} 
		   ~ & ~ & MedR$\downarrow$ & R@1$\uparrow$ &R@5$\uparrow$ & R@10$\uparrow$ & MedR$\downarrow$ & R@1$\uparrow$ &R@5$\uparrow$ & R@10$\uparrow$ \\   
		\hline
		\multirow{16} *{1k}& SAN~\cite{JinJinChen+MM2017_SAN}&16.1 & 12.5 & 31.1 & 42.3 & - & - & - & - \\ 
		~ & JESR ~\cite{Salvador+CVPR2017_JESR}  & 5.2 & 24.0 & 51.0 & 65.0 & 5.1 & 25.0 & 52.0 & 65.0 \\ 
		~ & Img2img+JESR~\cite{lien2020recipe}  & - & - & - & - & 5.1 & 23.9 & 51.3 & 64.1 \\ 
		~ & AMSR~\cite{JinJinChen+MM2018_AMSR}  & 4.6 & 25.6 & 53.7 & 66.9 & 4.6 & 25.7 & 53.9 & 67.1 \\ 
		~ & AdaMine~\cite{Carvalho+SIGIR2018_AdaMine}  & 1.0 & 39.8 & 69.0 & 77.7 & 1.0 & 40.2 & 68.1 & 78.7 \\ 
		~ & AdaMine$^*$~\cite{Carvalho+SIGIR2018_AdaMine}  & 2.3 & 36.7 & 67.0 & 76.8 & 2.2 & 38.0 & 67.6 & 77.3 \\ 
		~ & R$^2$GAN~\cite{zhu2019r2gan} &2.0 & 39.1 & 71.0 & 81.7 & 2.0 & 40.6 & 72.6 & 83.3 \\
		~ & MCEN~\cite{fu2020mcen} &2.0 & 48.2 & 75.8 & 83.6 & 1.9 & 48.4 & 76.1 & 83.7 \\
		~ & ACME~\cite{Hao+CVPR2019_ACME} &1.0 & 51.8 & 80.2 & 87.5 & 1.0 & 52.8 & 80.2 & 87.6 \\
		~ & ACME$^*$~\cite{Hao+CVPR2019_ACME} &1.4 & 50.0 & 78.9 & 86.4 & 1.3 & 51.1 & 79.1 & 86.5 \\
		 \cline{2-10} 
		~ & \textbf{SEJE}(TextRank, ResNet-50) & 1.0 & 51.9  & 81.5 & 88.9 & 1.0  & 53.0  & 82.1  & 89.1 \\ 
		~ & \textbf{SEJE}(DistilBERT, ResNet-50) & 1.0 & 52.3  & 81.6 & 88.6 & 1.0  & 53.6  & 82.0  & 89.2 \\ 
		~ & \textbf{SEJE}(BERT, ResNet-50) & 1.0 & 52.8  & 81.7 & 89.2 & 1.0  & 53.7  & 82.3  & 89.5 \\ 
		~ & \textbf{SEJE}(RoBERTa, ResNet-50) & 1.0 & 54.6  & 83.3 & 90.4 & 1.0  & 55.2  & 83.7  & 90.7 \\ 
		~ & \textbf{SEJE}(TF-IDF, ResNet-50) & 1.0 & 57.2  & 85.2 & 91.2 & 1.0  & 57.4  & 85.7  & 91.7 \\ 
		 \cline{2-10}
		 ~ & \textbf{SEJE}(RoBERTa, ResNeXt-101) & 1.0 & 56.3  & 85.1 & 91.8 & 1.0  & 57.6  & 85.6  & 92.1 \\ 
		~ & \textbf{SEJE(TF-IDF, ResNeXt-101)} & 1.0 & \textbf{58.1}  & \textbf{85.8} & \textbf{92.2} & 1.0  & \textbf{58.5}  & \textbf{86.2}  & \textbf{92.3}  \\ 
		\hline 
		 \multirow{15} *{10k} & JESR~\cite{Salvador+CVPR2017_JESR} & 41.9 & - & - & - & 39.2 & - & - & - \\ 
		~ & AMSR~\cite{JinJinChen+MM2018_AMSR}   & 39.8 & 7.2 & 19.2 & 27.6 & 38.1 & 7.0 & 19.4 & 27.8 \\ 
		~ & AdaMine~\cite{Carvalho+SIGIR2018_AdaMine}  & 13.2 & 14.9 & 35.3 & 45.2 & 12.2 & 14.8 & 34.6 & 46.1 \\ 
		~ & AdaMine$^*$~\cite{Carvalho+SIGIR2018_AdaMine}  & 16.4 & 12.8 & 31.5 & 42.2 & 15.5 & 13.7 & 32.9 & 43.6 \\ 
		~ & R$^2$GAN~\cite{zhu2019r2gan} &13.9 & 13.5 & 33.5 & 44.9 & 12.6 & 14.2 & 35.0 & 46.8 \\
		~ & MCEN~\cite{fu2020mcen} & 7.2 & 20.3 & 43.3 & 54.4 & 6.6 & 21.4 & 44.3 & 55.2 \\
		~ & ACME~\cite{Hao+CVPR2019_ACME}& 6.7 & 22.9 & 46.8 & 57.9 & 6.0 & 24.4 & 47.9 & 59.0 \\ 		
		~ & ACME$^*$~\cite{Hao+CVPR2019_ACME} &7.0 & 21.5 & 45.4 & 56.6 & 6.8 & 23.0 & 46.5 & 57.7 \\
		 \cline{2-10} 
		~ & \textbf{SEJE}(TextRank, ResNet-50) & 6.2 & 21.9  & 47.2 & 59.2 & 6.0  & 22.8  & 48.1  & 59.9 \\ 
		~ & \textbf{SEJE}(DistilBERT, ResNet-50) & 6.0 & 22.7  & 48.3 & 60.1 & 6.0  & 23.7  & 48.8 & 60.5 \\ 
		~ & \textbf{SEJE}(BERT, ResNet-50) & 6.0 & 23.2  & 48.8 & 60.6 & 6.0  & 24.0  & 49.6  & 61.2 \\ 
		~ & \textbf{SEJE}(RoBERTa, ResNet-50) & 5.1 & 24.0  & 50.3 & 62.2 & 5.0  & 24.9  & 50.8  & 62.6 \\
		~ & \textbf{SEJE}(TF-IDF, ResNet-50) & 4.9 & 26.3  & 53.2 & 64.9 & 4.8  & 27.0  & 53.7  & 65.3  \\ 
		\cline{2-10}
		 ~ & \textbf{SEJE}(RoBERTa, ResNeXt-101) & 5.0 & 25.6 & 52.3 & 64.4 & 4.9  & 26.6  & 53.3  & 65.1 \\ 
		~ & \textbf{SEJE(TF-IDF, ResNeXt-101)} & \textbf{4.2} & \textbf{26.9}  & \textbf{54.0} & \textbf{65.6} & \textbf{4.0}  & \textbf{27.2}  & \textbf{54.4}  & \textbf{66.1}  \\  
		\hline
		\end{tabular} 
\end{table}

\subsection{Baselines for Comparison.}  We consider the following state-of-the-art methods as the baselines in evaluation: 
 
\textbf{Stacked Attention Networks (SAN)~\cite{JinJinChen+MM2017_SAN}}: SAN applied a stacked attention network to simultaneously locate ingredient regions in the image and learn multi-modal embedding features between ingredient features and image features through a two-layer attention mechanism.

\textbf{Joint Embedding with Semantic Regularization (JESR) ~\cite{Salvador+CVPR2017_JESR}}: JESR obtains the recipe representation using LSTM networks on the ingredients and instructions text, the image representation using ResNet-50, and train a joint embedding of the two modalities using pairwise cosine loss with a regularization term to penalize the learned embedding features if they fail in performing food categorization.

\textbf{Recipe Retrieval with visual query of ingredients (Img2img+JESR) ~\cite{lien2020recipe}}: Based on the framework of JESR, Img2img+JESR appends the representation of the ingredient image set to the concatenation of ingredient embedding and instruction embedding, and then the concatenation of three embeddings will be transformed into one joint recipe embedding through a fully-connected layer.

\textbf{Attention Mechanism with Semantic Regularization (AMSR) ~\cite{JinJinChen+MM2018_AMSR}}: AMSR utilizes GRU networks and attention mechanism to encode recipe text at different levels (title, ingredients and instructions), and adopts the triplet loss rather than the pairwise loss to train the model.

\textbf{ADAptive MINing Embedding (AdaMine) ~\cite{Carvalho+SIGIR2018_AdaMine}}: AdaMine utilizes a double triplet loss where the two triplets are built on the matching recipe-image relationship and their category labels, with the adaptive learning integrated into their triplet loss strategy.

\textbf{Recipe Retrieval with GAN (R$^2$GAN)~\cite{zhu2019r2gan}}: R$^2$GAN is built upon the embedding learning framework of JESR, the generative adversarial network with multiple generators and discriminators and two-level ranking loss to learn compatible embeddings for cross-modal similarity measure.

\textbf{Modality-Consistent Embedding Network (MCEN)~\cite{fu2020mcen}}: MCEN exploits the latent alignment with cross-modal attention mechanism and shares the cross-modal information to the joint embedding space with stochastic latent variable models.

\textbf{Adversarial Cross-Modal Embedding (ACME)~\cite{Hao+CVPR2019_ACME}}: ACME uses the triplet loss empowered with hard sample mining and an adversarial loss to align the embeddings from the two modalities. Besides, they try to utilize the cross-modal translation consistency component to save the information lost in the training process.

\begin{figure}[t]
  \centering  
  \includegraphics[scale=0.21]{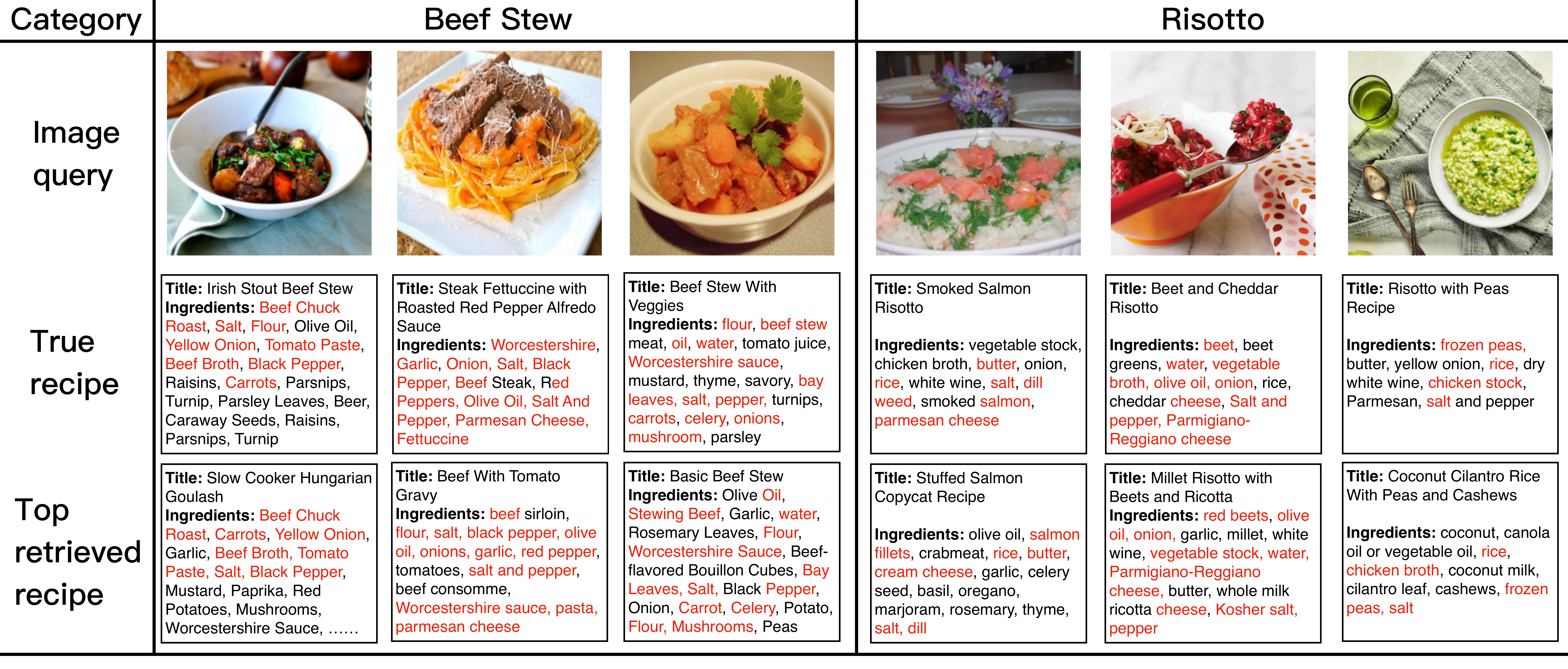}  
  \caption{The results of image-to-recipe retrieval by our SEJE approach on the 10k test set. The common or similar ingredients in the true recipe and top retrieved recipe are highlighted in red.}   
  \label{JEMA-i2r results}  
\end{figure} 
\subsection{Cross-Modal Retrieval Performance}
\noindent
We evaluate the performance of our SEJE approach for image-to-recipe and recipe-to-image retrieval against the six baselines in Table~\ref{main_results}. To provide a fair comparison, the results of SAN, JESR, Img2img+JESR, AMSR, AdaMine, R$^2$GAN, ACME and MCEN are quoted directly from the original results reported in the respective literature~\cite{Salvador+CVPR2017_JESR,lien2020recipe,JinJinChen+MM2018_AMSR,Carvalho+SIGIR2018_AdaMine,zhu2019r2gan,Hao+CVPR2019_ACME,fu2020mcen}.

%And as pointed out in~\cite{zhu2019r2gan}, the ranking position starts from 0 instead of 1 when calculating the evaluation metrics in AdaMine's implementation\footnote{https://github.com/Cadene/recipe1m.bootstrap.pytorch}, which is different from the calculation in other baselines. For the fair comparison, we evaluate the AdaMine approach on our testing set by using their provided pre-trained embeddings\footnote{http://data.lip6.fr/cadene/im2recipe/logs/adamine.tar.gz} (denoted as AdaMine$^*$). Since the state-of-the-art approach ACME releases the pre-trained model\footnote{https://drive.google.com/drive/folders/1svtpy-sD4pcaFfLGQNGaPIVjrKr-lhsT?usp=sharing}, we also evaluate their model on the same sampled testing subsets as ours (denoted as ACME$^*$). 
From the results in Table~\ref{main_results}, we observe that SEJE consistently outperforms all baselines with high Recall@K (K=1,5,10) for both image-to-recipe and recipe-to-image queries on 1k and 10k test data, showing the effectiveness of using the deep feature engineering techniques to incorporate the additional semantics for further optimizing the cross-modal joint embedding learning. And the attention mechanism introduced in SAN and AMSR on top of the neural features aims to approximately locate the key ingredients. However, the results are not effective, especially compared to SEJE which combines the term rating module and image category for optimizing the joint embedding learning.

\begin{figure}[t]
  \centering  
  \includegraphics[scale=0.95]{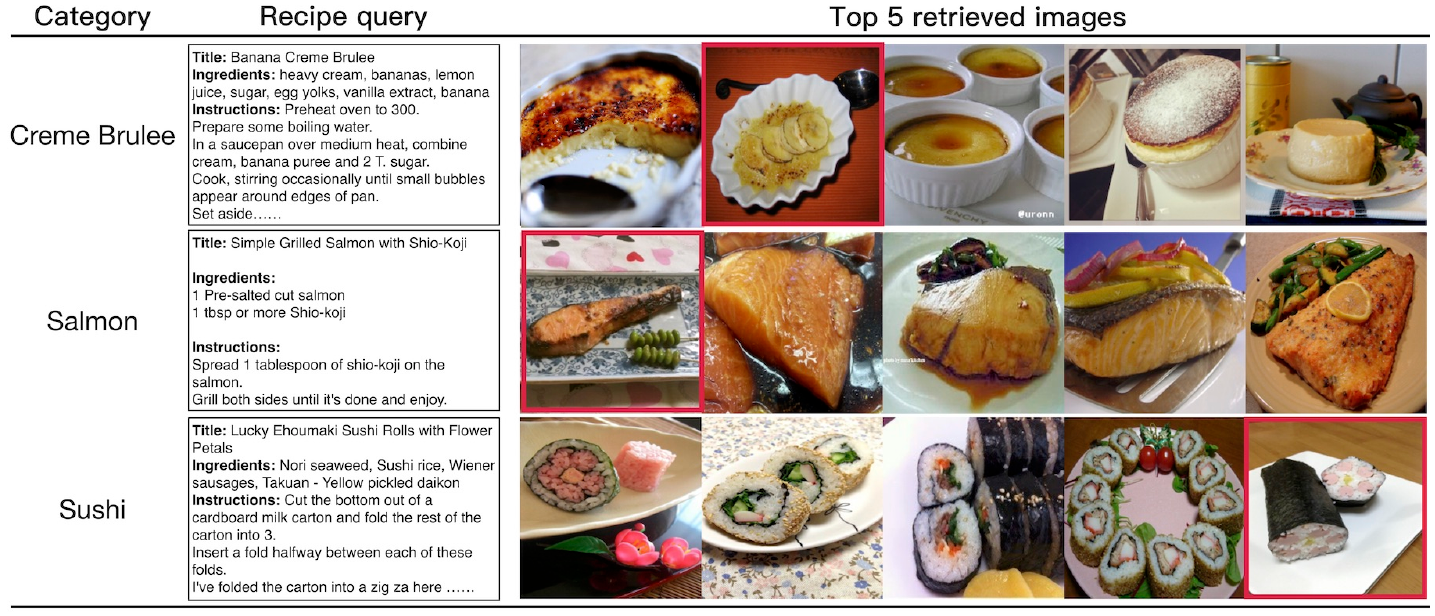}  
  \caption{The results of recipe-to-image retrieval by our SEJE approach on the 10k test set. The matched images are boxed in red.}   
  \label{JEMA-r2i results}   
\end{figure} 

We also evaluate different term rating algorithms used in the term rating module: TF-IDF, TextRank and three popular BERT models, which are standard BERT~\cite{devlin2018_bert}, DistilBERT~\cite{sanh2019_distilbert} and RoBERTa~\cite{liu2019_roberta}. The results are reported in Table~\ref{main_results} under SEJE sections, which indicates that selecting the TF-IDF approach in the term rating module can yield the best accuracy in SEJE, and RoBERTa comes the second. We also vary the recent CNN models in the image embedding process by changing ResNet-50 to ResNet-152~\cite{He+CVPR2016}, WideResNet-101~\cite{wideResnet}, ResNeXt-50 and ResNeXt-101~\cite{xie2017aggregated}. It turns out that using ResNeXt-101 as the image encoder can yield stable improvement for all five term ranking algorithms, and TF-IDF remains to be the best in the context of SEJE.

\begin{figure}[t]
  \centering
  \includegraphics[scale=0.22]{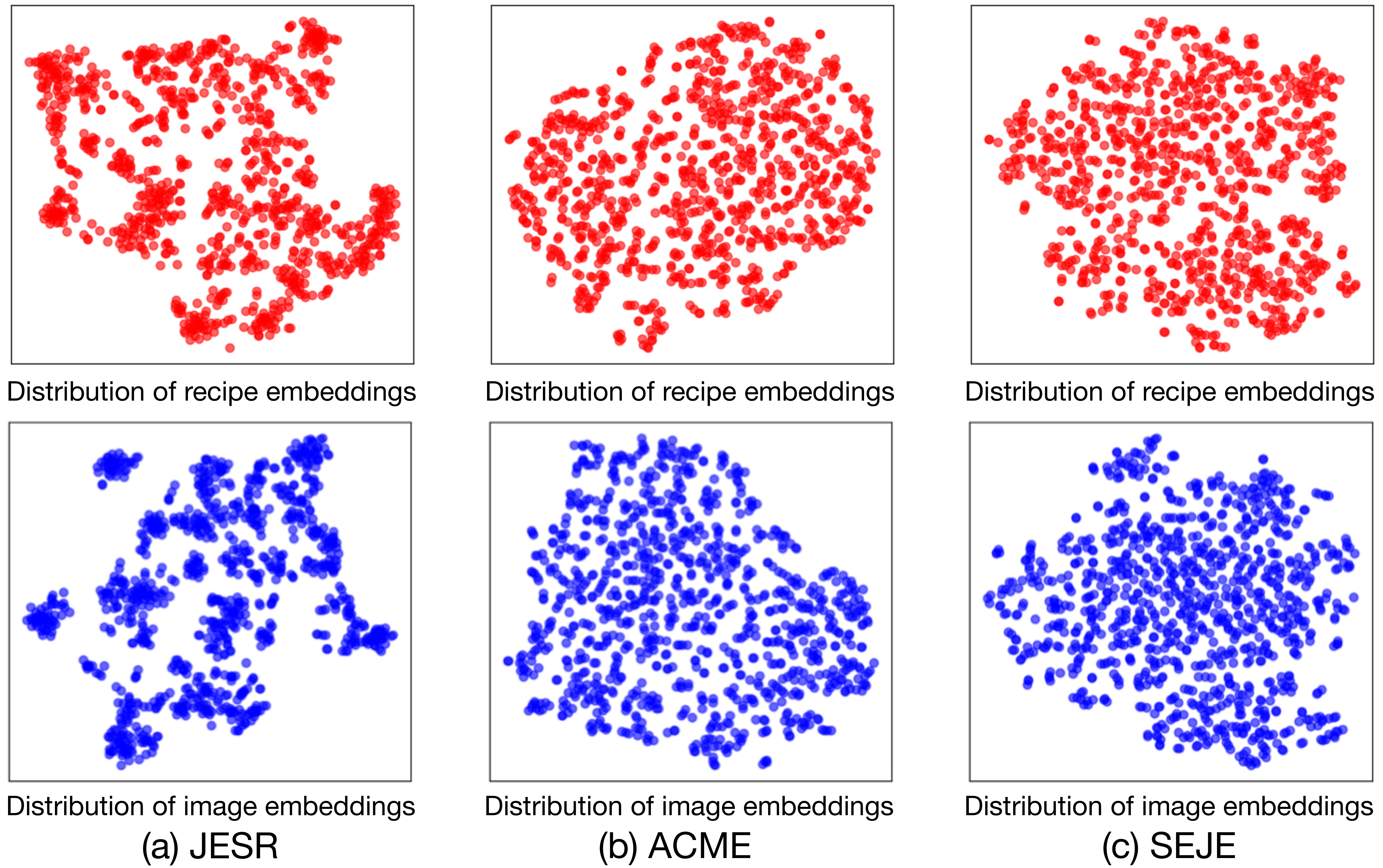}
  \caption{Comparing distributions of learned recipe and image embeddings by SEJE approach (c) with the distributions by JESR approach (a) and ACME approach (b), based on the t-SNE visualization on the 1000 randomly selected recipe-image pairs from the test set of Recipe1M dataset. }
  \label{t-sne}
\end{figure}

\subsection{Visualization Results}
\noindent We provide some visualization results of SEJE for both image-to-recipe and recipe-to-image retrieval tasks on the 10k dataset in Figure~\ref{JEMA-i2r results} and Figure~\ref{JEMA-r2i results} respectively. 
Figure~\ref{JEMA-i2r results} shows the ground truth recipes and the top retrieved recipes based on 6 different image queries. We consider two recipe categories: \textit{beef stew} and \textit{risotto}, and in each category we list three image queries. It shows that almost all the significant ingredients of the true recipes and visual components in the image queries also appear in the top retrieved recipes. Figure~\ref{JEMA-r2i results} visualizes the results of retrieving the top 5 images using three different recipe queries. In all cases, most of the retrieved images share similar ingredients to the ground truth images and contain the visual components as suggested by the category label.
In the first example, even though the background and decorations in the retrieved food images are quite different, all the retrieved top 5 images can be easily recognized as creme brulee, as suggested by the category label. In the second example, all the top 5 results contain the main ingredient ``salmon'' and the matched image is retrieved as top 1. All the retrieved images in the third example are visually similar and all are about sushi with a petal design. 
Figure~\ref{JEMA-r2i results} and Figure~\ref{JEMA-i2r results} further illustrate the effectiveness of our SEJE approach for text-image joint embedding learning and the two-phase deep feature engineering can boost the accuracy of the cross-modal retrieval task.

\begin{figure} [t] 
  \centering   
  \includegraphics[scale=0.95]{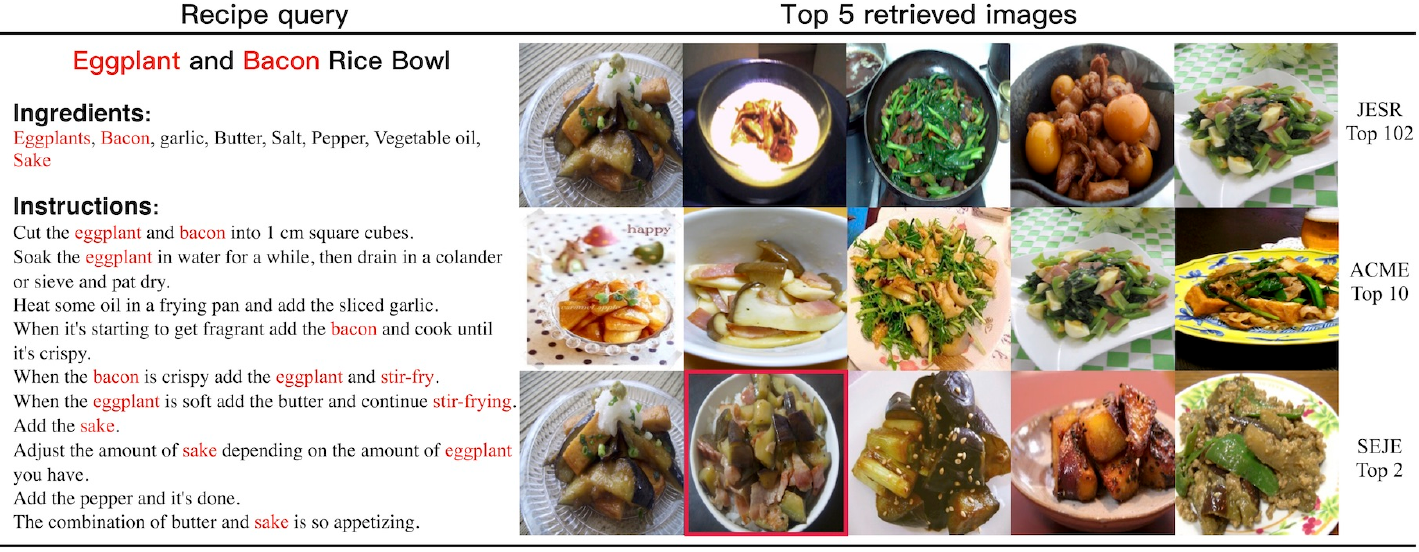}   
  \caption{Comparing our SEJE approach with two representative methods JESR~\cite{Salvador+CVPR2017_JESR} and ACME~\cite{Hao+CVPR2019_ACME} on the recipe-to-image retrieval (10K test set). The matched images are marked in the red box. Words in red highlighted in recipe text indicate that they are selected with relatively high TF-IDF value in our SEJE. We indicate the top-$k$ position where the matched image is retrieved under each of the three methods.}    
  \label{main-r2i results}   
\end{figure} 

To show the effectiveness and high quality of our SEJE learned joint embeddings, the performance of our SEJE approach is visualized to compare with the existing representative approaches. Since JESR~\cite{Salvador+CVPR2017_JESR} and ACME~\cite{Hao+CVPR2019_ACME} release their pre-trained models for the recipe-text cross-modal retrieval task, here we will compare our SEJE with these two approaches. Figure~\ref{t-sne} shows the distributions of the learned recipe and image embeddings by the JESR, ACME and our SEJE approaches, also based on the t-SNE visualization on 1000 randomly selected recipe-image pairs from the test set of Recipe1M dataset. We can find that the distributions of recipe and image embeddings learned by JESR vary greatly, which accounts for the low accuracy results in the cross-modal retrieval tasks. The distributions learned by ACME are much better but still not similar. Our SEJE approach aligns the distributions of learned recipe and image embeddings well and their distributions are visually similar.

\begin{figure}
  \centering
  \includegraphics[scale=0.2]{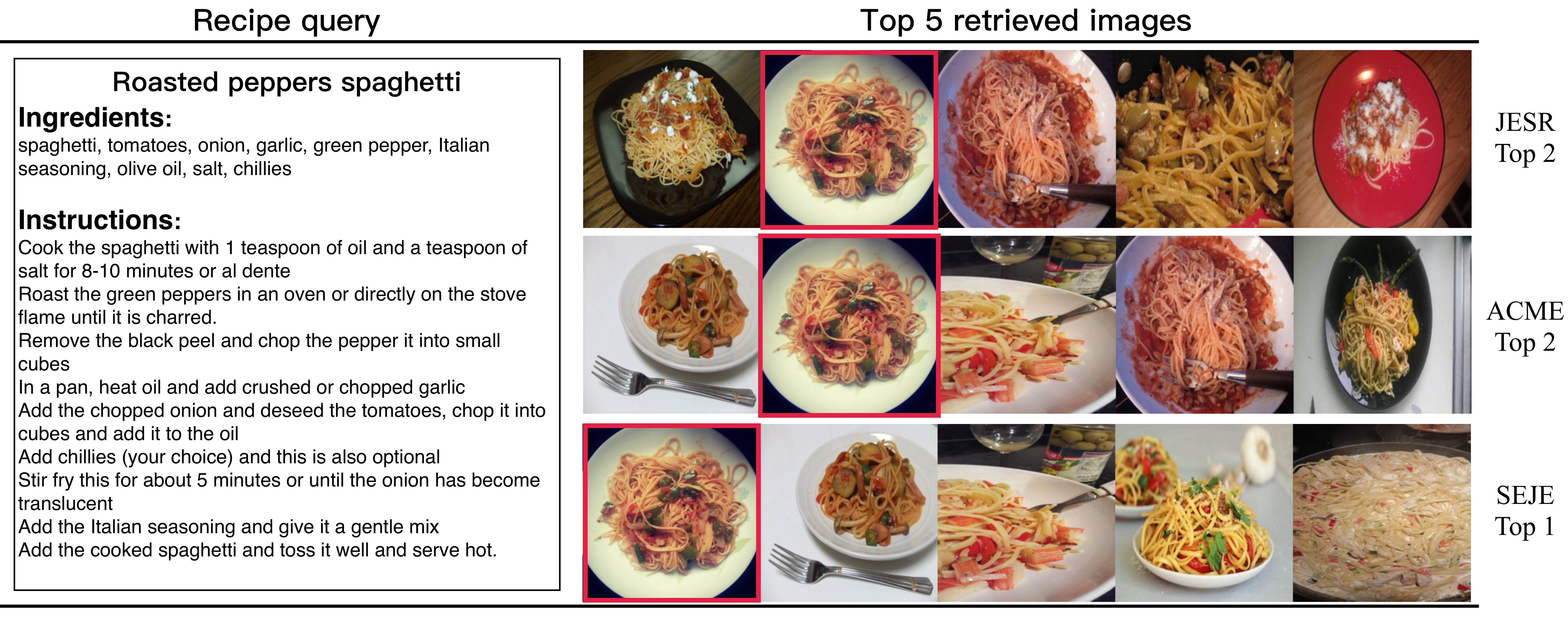}
  \caption{An easy-case example of recipe-to-image retrieval, in which all three approaches (JESR, ACME, and our SEJE) perform well. The matched images are highlighted in a red box. We also list the top-k positions where the matched images are retrieved.}
  \label{easy}
\end{figure}

\begin{figure}
\centering
\includegraphics[scale=0.2]{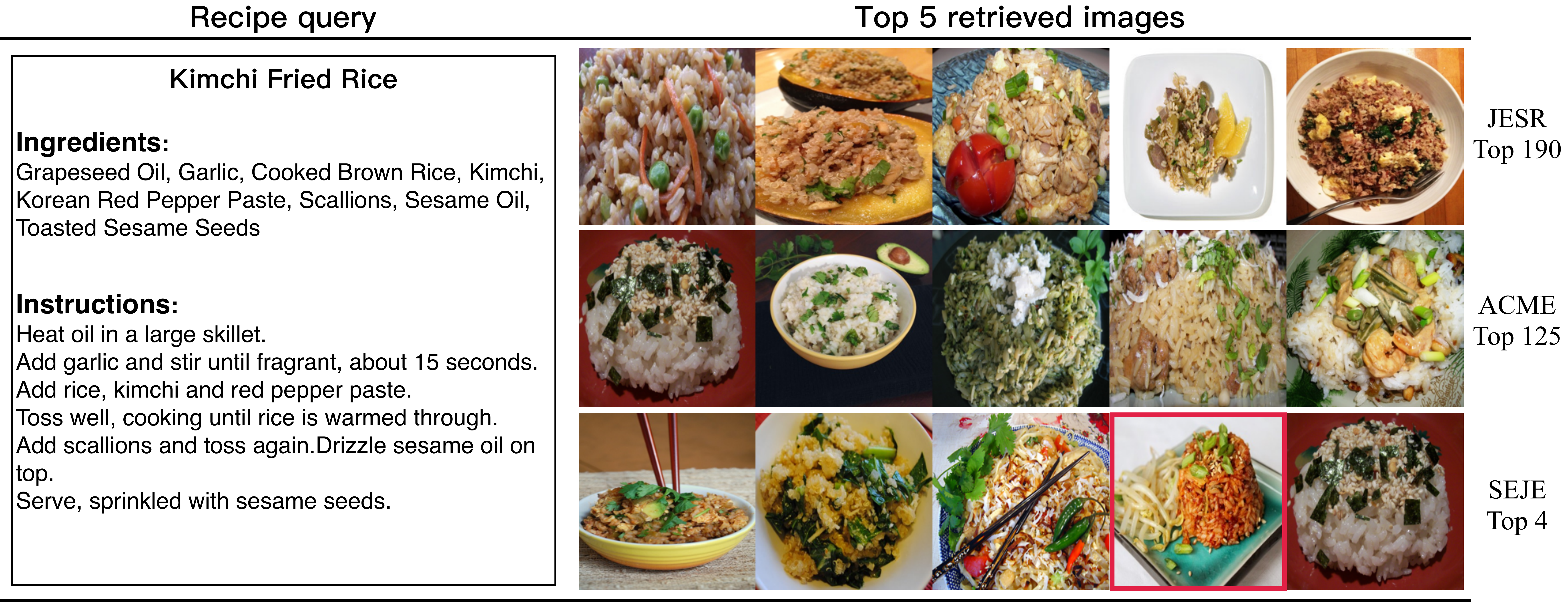}
\caption{An example of recipe-to-image retrieval, which illustrates the reasons that our SEJE approach outperforms JESR and ACME.  The matched images are highlighted in a red box. We also list the top-k positions where the matched images are retrieved. By leveraging our key term extraction and ranking in Phase I feature engineering preprocessing, Kimchi is identified as the most important key ingredient in SEJE unlike  JESR and ACME.}
\label{average}
\end{figure}

To further illustrate the comparison results, we provide a
visualization of recipe-to-image retrieval using an example recipe by comparing the performance of SEJE with JESR and ACME in Figure~\ref{main-r2i results}, since JESR and ACME have released their pre-trained models. 
We observe that SEJE successfully aligns the ``eggplant'' ingredient in the recipe text with the eggplant component in the top retrieved food image and the matched image is returned at Top 2 by SEJE. In comparison, both JESR and ACME fail to retrieve the matched image within the top 5 results and most of the retrieved food images do not contain the eggplant component, showing some problems in modality alignment between recipe text and food image. This example further illustrates the effectiveness of our two-phase deep feature engineering framework for improving the quality of cross-modal retrieval tasks.

\begin{figure}
  \centering
  \includegraphics[scale=0.18]{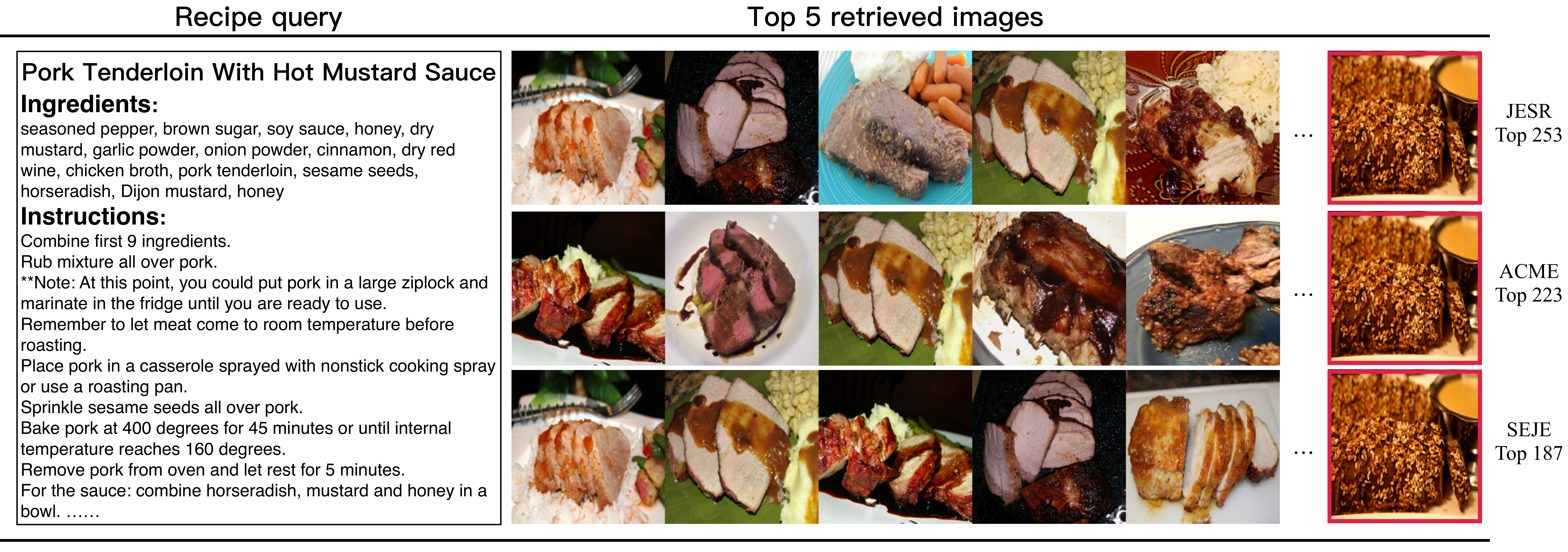}
  \caption{A hard-case example of recipe-to-image retrieval. All three approaches (JESR, ACME and our SEJE) perform poorly. The matched images are highlighted in a red box. We also list the top-k positions where the matched images are retrieved.}
  \label{hard}
\end{figure}

And here we provide several visualization-based analyses on the 10k test set to provide a more intuitive understanding of our SEJE approach in terms of easy, average and hard cases. Figure~\ref{easy} shows a recipe-to-image retrieval example where all three approaches return the matched image within the top 5 results. This is a representative example of easy scenarios. Figure~\ref{average} shows a recipe-to-image retrieval example where SEJE returns the matched image within the top 5 results whereas JESR and ACME return the matched image at top 190 and top 125 respectively. Figure~\ref{hard} shows an example of recipe-to-image retrieval that represents the most challenging scenarios. Here not only all three approaches fail to return the matched result in the top 5 but also the matched result was returned at top 253, top 223 and top 187 for JESR, ACME and our SEJE respectively. The recipe query is about the pork tenderloin, but the matched food image is hard for even a human to recognize and it can also visually be mistaken as sliced bread. That's why all the approaches fail in this case. But the return images of the three approaches mostly are about the pork tenderloin, which are still acceptable as the returned results.

\begin{figure}
  \centering   
  \includegraphics[scale=0.6]{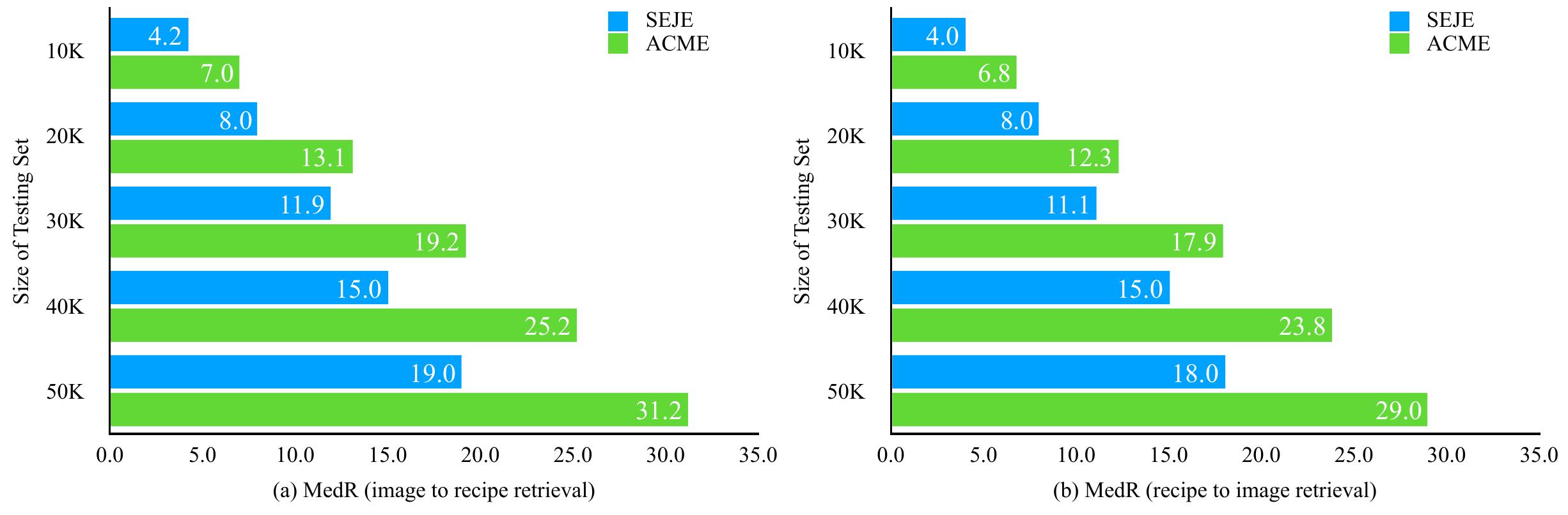}   
  \caption{ Scalability test between SEJE and ACME for both image-to-recipe retrieval and recipe to image retrieval (For MedR, lower is better).} 
  \label{MedR}   
\end{figure}

{\bf Scalability.} To investigate the robustness of SEJE against large datasets beyond 10K, we further compare its MedR performance against the state-of-the-art approach ACME. Figure~\ref{MedR} shows the results for image-to-recipe retrieval and recipe-to-image retrieval. The gap between our SEJE and ACME becomes larger as the testing set size increases. SEJE successfully ranks the ground-truth recipes by over 12.2 positions ahead of ACME on average on the 50K testing set, which is almost equivalent to the original complete testing set~\cite{Salvador+CVPR2017_JESR}. Similar results are also observed for the recipe-to-image retrieval, where SEJE outperforms ACME by over 11 positions on average for the 50K dataset. 

\begin{table}[t]
		\center
		\caption{Evaluation of contributions of different components in the SEJE framework for the image-to-recipe retrieval on the 1k test-set.
}
		\label{ablation results}
		\begin{tabular}{c|ccccc|cccc|cccc}
		\hline
		 \multicolumn{6}{c|}{Component} & \multicolumn{4}{c|}{Image to recipe retrieval } & \multicolumn{4}{c}{Recipe to image retrieval } \\
		\hline
		 SEJE-b & En$_{V}$ & En$_{R}$ & CA & DA & TRI &MedR & R@1 &R@5 & R@10 & MedR & R@1 &R@5 & R@10 \\  
		\hline
		$\checkmark$ & ~ & ~ & ~ & ~ & ~ & 4.1 & 25.9 & 56.4 & 70.1 & 4.1 & 26.0 & 56.6 & 70.3\\
		$\checkmark$& $\checkmark$ & ~ & ~ & ~ & ~ & 3.4 & 28.1 & 59.5 & 73.1 & 3.0 & 29.4 & 61.2 & 74.4  \\
		$\checkmark$& ~ & $\checkmark$ & ~ & ~ & ~ & 3.0 & 29.4 & 60.0 & 73.4 & 3.0 & 31.0 & 61.0 & 73.9 \\		
		$\checkmark$& $\checkmark$ & $\checkmark$ & ~ & ~ & ~ & 3.0 & 30.5 & 61.6 & 75.2 & 2.8 & 32.1 & 62.6 & 75.4\\
		\hline
		$\checkmark$& $\checkmark$ & $\checkmark$ & $\checkmark$ & ~ & ~ & 2.3 & 33.5 & 68.4 & 80.9 & 2.3 & 34.1 & 68.3 & 81.4\\
		$\checkmark$& $\checkmark$ & $\checkmark$ & ~ & $\checkmark$ & ~ &  2.5 & 36.0 & 65.2 & 77.3 & 2.5 & 36.0 & 65.2 & 77.8\\
		$\checkmark$& $\checkmark$ & $\checkmark$ & ~ & ~ & $\checkmark$ &  1.6 & 47.7 & 78.6 & 87.3 & 1.6 & 48.4 & 79.1 & 87.5 \\ 
		$\checkmark$& $\checkmark$ & $\checkmark$ & $\checkmark$ & $\checkmark$ & $\checkmark$ & \textbf{1.0}  & \textbf{57.2}  & \textbf{85.2} & \textbf{91.2}& \textbf{1.0}  & \textbf{57.4}  & \textbf{85.7}  & \textbf{91.7}\\  
		\hline
		\end{tabular}
\end{table}

\subsection{Ablation Study}

\noindent
This section studies the improvements owing to each key component in our two-phase deep feature engineering enhanced approach for image-to-recipe and recipe-to-image retrieval tasks. For a fair comparison with the baselines, ResNet-50 model and TF-IDF approach are set as default in the image embedding learning and term rating module for this set of experiments. Table~\ref{ablation results} reports the results. \textbf{SEJE-b} denotes the basic SEJE framework with the batch-all triplet loss and without any of our proposed deep feature engineering techniques. We then incrementally add one component at a time. First, we investigate the significance of employing the deep feature engineering for the input image and recipe preprocessing to leverage the additional image and recipe semantics respectively in the modality-specific embedding learning, i.e., \textbf{En$_V$} and \textbf{En$_R$}. For the Phase II enhanced joint embedding learning, we first analyze the gains of the category-based alignment loss regularizations \textbf{CA} on both recipe and image embeddings. Next, we add the cross-modal discriminator-based alignment loss regularization \textbf{DA}. Then we use our double negative sampling and soft-margin optimized batch-hard triplet loss (denoted as \textbf{TRI}) to replace the batch-all triplet loss in \textbf{SEJE-b}. Table~\ref{ablation results} shows that each proposed component positively contributes towards improving the cross-modal alignment between recipe text and food image, effectively boosting the overall performance of cross-modal retrieval.

\begin{table} [t]
		\center 
		\caption{ Evaluation of contributions of each component in the extracted key terms on the 10K test-set.} 
		\label{term component} 
		\begin{tabular}{c|cccc|cccc} 
		\hline
		 \multirow{2} * {Components}  &\multicolumn{4}{c}{Image to recipe retrieval} & \multicolumn{4}{c}{Recipe to image retrieval } \\ 
		 \cline{2-9} 
		   ~ & MedR$\downarrow$ & R@1$\uparrow$ &R@5$\uparrow$ & R@10$\uparrow$ & MedR$\downarrow$ & R@1$\uparrow$ &R@5$\uparrow$ & R@10$\uparrow$ \\   
		\hline
		 ingredient & 5.0 & 25.0  & 51.4 & 63.1 & 5.0  & 26.1  & 52.0  & 63.4 \\
		 ingredient+utensil & 5.0 & 25.6  & 52.2 & 64.1 & 5.0  & 26.4  & 52.5  & 64.2 \\ 
		 ingredient+utensil+action & 5.0  & \textbf{26.3}  & \textbf{53.2} & \textbf{64.9} & \textbf{4.9}  & \textbf{27.0}  & \textbf{53.7}  & \textbf{65.3} \\ 
		\hline 
		\end{tabular} 
\end{table}

\begin{table} [t]
		\center 
		\caption{ Performance comparison of our SEJE using different key term filters on the 10K test set. ResNet-50 model is used here as the image encoder. The symbol ``-'' indicates that all the extracted key terms will be used to generate the key term feature instead of only using the filtered terms. }
		\label{weight threshold} 
		\begin{tabular}{cc|cccc|cccc} 
		\hline
		\multirow{2} * {\tabincell{c}{Term Ranking \\Algorithm}} & \multirow{2} * {Threshold}  &\multicolumn{4}{c}{Image to recipe retrieval} & \multicolumn{4}{c}{Recipe to image retrieval } \\ 
		 \cline{3-10} 
		   ~ & ~ & MedR$\downarrow$ & R@1$\uparrow$ &R@5$\uparrow$ & R@10$\uparrow$ & MedR$\downarrow$ & R@1$\uparrow$ &R@5$\uparrow$ & R@10$\uparrow$ \\   
		\hline
		\multirow{4} *{RoBERTa} & - & 5.0 & 24.0  & 50.3 & 62.2 & 5.0  & 24.9  & 50.8  & 62.6 \\
		~ & 0.05 & 5.0 & 25.1  & 51.7 & 63.5 & 5.0  & 25.8  & 52.3  & 64.0 \\ 
		~ & 0.10 & 5.0 & 24.2  & 50.4 & 62.5 & 5.0  & 25.3  & 51.2  & 62.7 \\ 
		~ & 0.15 & 5.0 & 24.0  & 50.3 & 62.0 & 5.0  & 25.0  & 50.8  & 62.6 \\ 
		\hline 
		 \multirow{4} *{TF-IDF} & - & 5.0  & \textbf{26.3}  & \textbf{53.2} & \textbf{64.9} & \textbf{4.9}  & \textbf{27.0}  & \textbf{53.7}  & \textbf{65.3}  \\ 
		~ & 0.05 & 5.0 & 26.1  & 53.1 & 64.8 & 5.0  & 26.7  & \textbf{53.7} & 65.2 \\ 
		~ & 0.10 & 5.0 & 25.3  & 51.7 & 63.4 & 5.0  & 26.1  & 52.5  & 64.1 \\ 
		~ & 0.15 & 6.0 & 22.7  & 48.0 & 59.9 & 6.0  & 23.5  & 49.1  & 60.4 \\
		\hline
		\end{tabular} 
\end{table}

\begin{figure}
  \centering
  \includegraphics[scale=0.94]{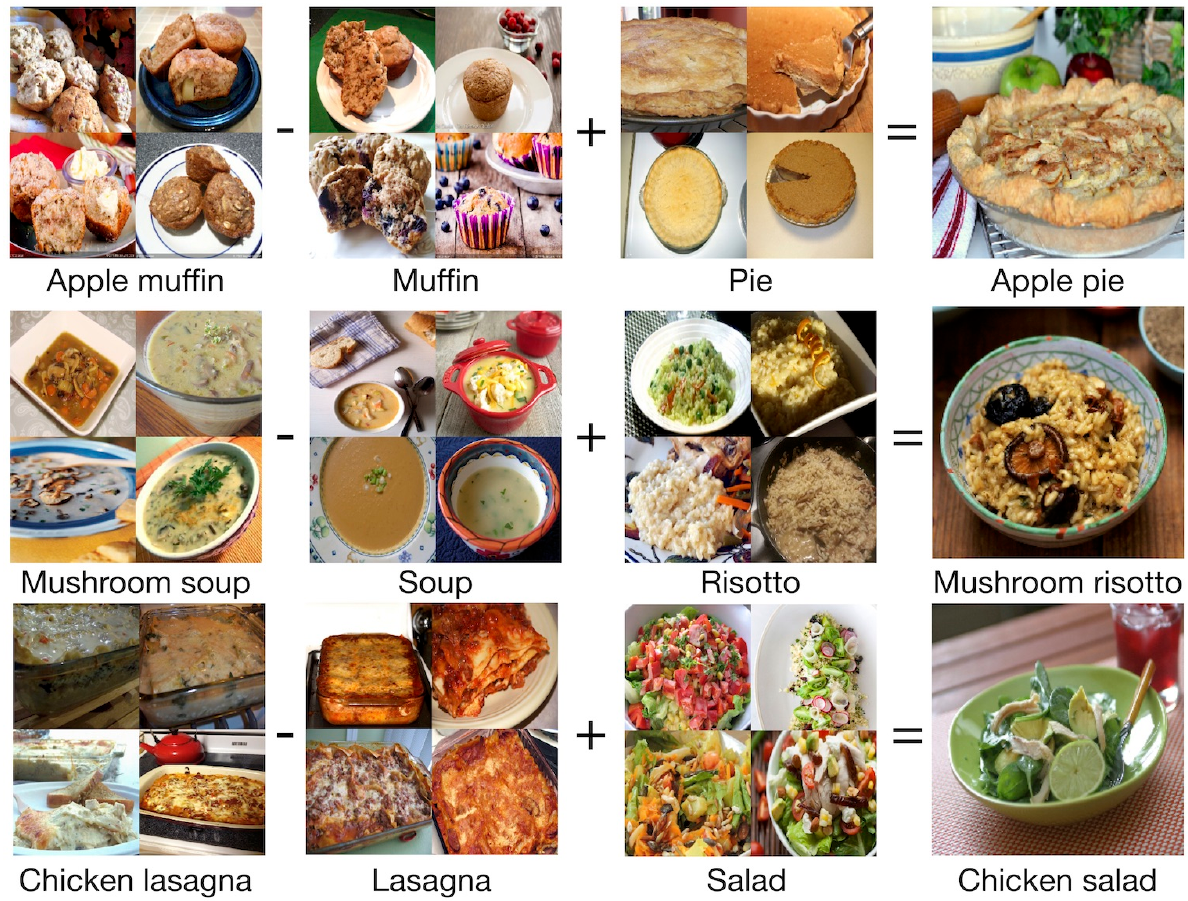}
  \caption{Semantic vector arithmetic results by using learned recipe embeddings on the Recipe1M test set. We represent the images associated with 4 nearest neighbor recipes concerning the query of the average vector of recipe embeddings whose recipe titles contain the keyword. For the arithmetic results, we only show the image associated with the nearest neighbor recipe.}
  \label{arithmetic_rec}
\end{figure}

\begin{figure}
  \centering
  \includegraphics[scale=0.94]{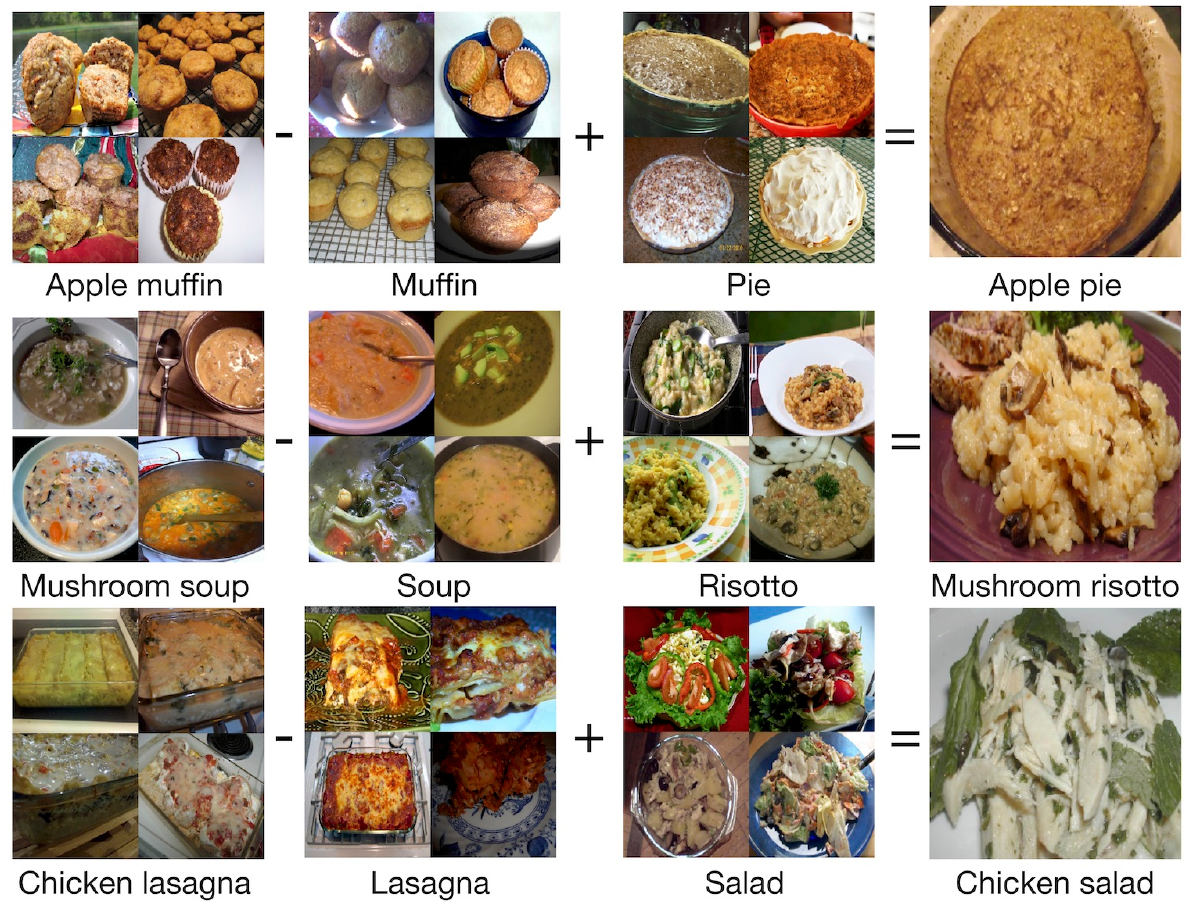}
  \caption{Semantic vector arithmetic results by using learned image embeddings on the Recipe1M test set. We represent the 4 nearest neighbor images concerning the query of the average vector of image embeddings whose corresponding recipes contain the keyword in the title. For the arithmetic results, we only show the nearest neighbor image.}
  \label{arithmetic_img}
\end{figure}

\begin{figure}
  \centering
  \includegraphics[scale=0.935]{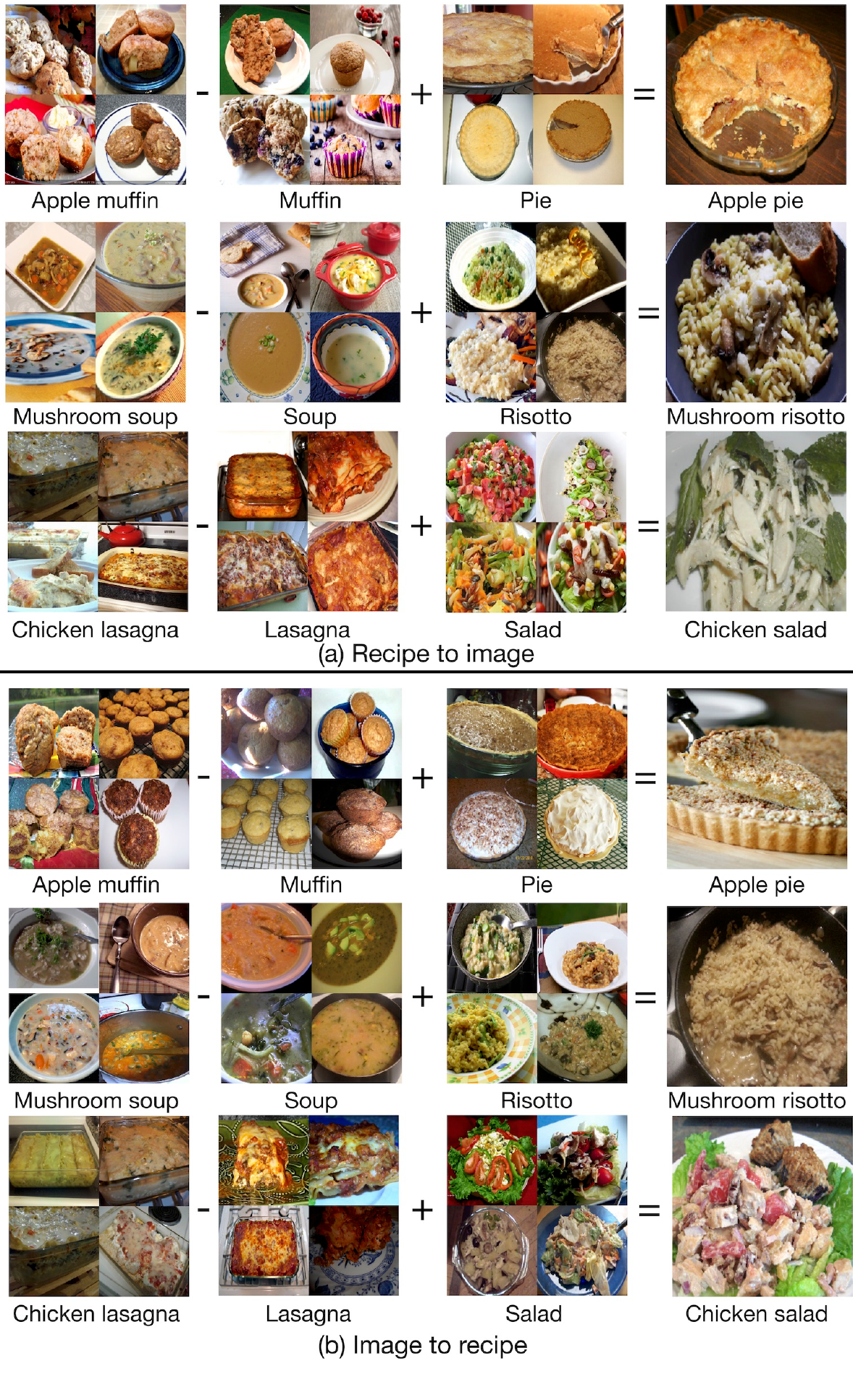}
  \caption{Cross-modal semantic vector arithmetic results by using learned recipe and image embeddings on the Recipe1M test set. We represent the 4 nearest neighbor images concerning the query of the average vector of embeddings whose recipe titles contain the keyword. For the arithmetic results, we only show the nearest neighbor image.}
  \label{arithmetic_cross}
\end{figure}

\subsection{ Effect of Term Extraction and Ranking}
As claimed in the introduction of the key term extraction module, the extracted key terms from the recipe text during the data preprocessing can be broadly divided into three types: ingredient entities, cooking utensils and actions. In this section, we want to investigate the significance of employing term extraction and term rating score on each of three types of key terms on the overall performance of SEJE. 

{\bf Evaluation of Each Key Term Component.\/} For the fair comparison, TF-IDF approach is set as the default choice in the term ranking module and ResNet-50 model is used in image embedding learning. First, we only extract the ingredient entities in our SEJE approach. Then we incrementally add the other two components: the cooking utensils and the cooking actions, each at a time. Table~\ref{term component} reports the results. It shows that every component positively contributes towards boosting the overall performance of cross-modal retrieval. 

{\bf Evaluation of the Key Term Filter.\/} 
In SEJE, each key term is assigned a different weight using the chosen term extraction and term ranking algorithm, such as TextRank, BERT based approach or TF-IDF, in order to capture the different levels of discrimination significance contributed by each key term to its recipe text compared to the other recipe texts in the entire Recipe1M dataset in the preprocessing phase I for deep feature engineering.
The more unique the key term is for its respective recipe, the higher rating score it should get. 
The SEJE approach by default leverages all the key terms extracted from the recipe text without filtering out those terms with very low weight. In the next set of experiments, we compare our SEJE default approach with the SEJE variants, which utilize a rating threshold based filter to remove those terms whose ranking scores (weights) are lower than the given threshold $t$ when generating the weighted key term feature for recipe text. Table~\ref{weight threshold} reports the results of comparing TF-IDF with RoBERTa, a BERT variant for this set of experiments, since these two methods outperform other term rating algorithms (recall Table~\ref{main_results}). We observe that the RoBERTa based term rating can achieve higher accuracy and hence benefit from using the threshold based term filter when the rating threshold is set to 0.05. In comparison, the TF-IDF based term rating works consistently better for cross-modal retrieval performance compared to using TF-IDF variants using different threshold settings. This is another motivation for SEJE to take TF-IDF as the default NLP based term extraction and ranking method for better performance stability.

However, the importance of leveraging the terms with very low weight is not clear, which can be not helpful or even harmful to the quality of the learned recipe embedding. In order to investigate the impact of terms with low weight, we compare the performance of the SEJE variants that utilize a rating threshold based filter to remove the terms whose weights are lower than the threshold $t$ when generating the recipe weighted term feature, with the performance of our default SEJE version, which leverages all the key terms extracted from the recipe text. TF-IDF and RoBERTa-based approach are utilized in this set of experiments, since they outperform other term rating algorithms (recall Table~\ref{main_results}). Table~\ref{weight threshold} reports the results, which indicate that using the term filter can benefit the performance of using the RoBERTa-based term rating algorithm and better accuracy can be achieved when the rating threshold is set as 0.05, while no-filter on TF-IDF based approach offers consistently better cross-modal retrieval performance 
compared to different threshold settings, showing another benefit of the TF-IDF based term rating algorithm to the performance stability of SEJE.

\section{ SEJE Cross-Modal Embeddings for Semantic Vector Arithmetic}
One of the important measures for cross-modal embeddings is to measure the capabilities of their learned representations using simple arithmetic operations~\cite{mikolov+2013,radford+2015,Salvador+CVPR2017_JESR,marin+2019}. For the context of cooking recipes and food images, it would be expected that $v$(``Apple muffin'')-$v$(``Muffin'')+$v$(``Pie'')=$v$(``Apple pie''), where $v$($\cdot$) represents the feature embedding with the keyword in our learned joint embedding space. Specifically, given a recipe of an apple muffin, we can transform the learned embedding of apple muffin into an apple pie by subtracting and adding the mean recipe embeddings of muffin and pie respectively. We apply this procedure in the recipe and image embedding spaces by using the previous equation template to the averaged vectors whose corresponding recipes contain the queried words in their title. Figure~\ref{arithmetic_rec} and Figure~\ref{arithmetic_img} show the results in the recipe and image embedding spaces respectively. 
Furthermore, we apply the same arithmetic operation to the embeddings across modalities. Figure~\ref{arithmetic_cross} reports the results. These semantic vector arithmetic results indicate that the recipe and image embeddings learned in our SEJE model are semantically well-aligned, which facilitates the future works of applications in recipe ingredient analysis and cross-modal generation. 

%\section{Discussion}
%In what follows, we discuss the potential application tasks for our cross-modal retrieval model. Nowadays, there are a lot of public unlabeled, multi-modal datasets, which can be found from tweets, Yelp restaurant/dish recommendations, Instagram, medical imaging reports repositories, and so on. For example, by using the LIDC-IDRI dataset\footnote{https://wiki.cancerimagingarchive.net/display/Public/LIDC-IDRI}, a medical researcher wishes to retrieve relevant diagnostic text reports when given a lung cancer screening thoracic computed tomography (CT) scan image, and similarly, a doctor may wish to retrieve all relevant CT scan images from the repository when given a diagnostic report. Such multi-modal query services are also common among social media users from Twitter, Yelp, Instagram. By incorporating our two-phase deep calibration techniques for boosting the performance of cross-modal retrieval, many real world applications can benefit from it to improve their service and enhance their market competitiveness of customers and forward-looking, such as Yelp, Amazon, Tiktok and Instagram. 

\section{Conclusion}

We have presented SEJE, a two-phase deep feature engineering framework for learning cross-modal joint embedding with the semantics enhanced feature embeddings and loss optimizations. Our SEJE method can extract and incorporate the additional semantics in both cooking recipe text and food image to capture more discriminative properties of input recipe text and its associated food image, and semantically align the learned recipe and image embeddings. By further integrating with the double negative sampling and soft-margin optimized batch-hard triplet loss as the primary loss and the category-based alignment loss and discriminative-based alignment loss as the two auxiliary loss regularizations, SEJE effectively boosts the accuracy and retrieval performance of cross-modal joint embedding learning and it outperforms the six representative state-of-the-art methods for both image-to-recipe and recipe-to-image retrieval on Recipe1M benchmark dataset.   

Our research on cross-modal retrieval continues along two directions. First, we are interested in extending our SEJE deep feature engineering framework to cross-modal retrieval in a medical repository, e.g., using the LIDC-IDRI dataset\footnote{https://wiki.cancerimagingarchive.net/display/Public/LIDC-IDRI}, given that a medical repository with a medical document with associated medical imaging is semi-structured with the data structure sharing some similarity with the RecipeIM dataset. For example, one can find relevant diagnostic text reports when given a lung cancer screening thoracic computed tomography (CT) scan image. Similarly, one can retrieve all relevant CT scan images from the medical repository when given a diagnostic report. 
Second, we are interested to further extend our SEJE framework to less structured or unstructured datasets, such as social media like Twitter, Instagram, Tiktok, which can benefit from efficient cross-modal retrieval services.

\begin{acks}
We thank the EIC and the AE handling our paper and the reviewers for their constructive comments and suggestions. 
This work is partially supported by the USA National Science Foundation under Grants 2038029, 2026945, 1564097,  and an IBM faculty award. The first author has performed this work as a two-year visiting PhD student at Georgia Institute of Technology (2019-2021) with the support from China Scholarship Council (CSC) and Wuhan University of Technology.

\end{acks}

\bibliographystyle{ACM-Reference-Format}
\bibliography{TOIS}

\end{document}